\documentclass{CUP-JNL-EDS}

\usepackage{latexsym}
\usepackage{graphicx}
\usepackage{multicol,multirow}
\usepackage{amsmath,amssymb,amsfonts}
\usepackage{mathrsfs}
\usepackage{amsthm}
\usepackage{rotating}
\usepackage{appendix}
\usepackage[authoryear]{natbib}
\usepackage{ifpdf}
\usepackage[T1]{fontenc}
\usepackage{times}
\usepackage{sourcesanspro}
\usepackage{textcomp}%
\usepackage{xcolor}%
\usepackage{hyperref}
\usepackage{lipsum}

\usepackage{mathtools}
\usepackage{bm}
\usepackage{graphicx}
\usepackage{subcaption}
\usepackage{float}
\usepackage{hyperref}
\usepackage{xcolor, soul}
\usepackage{natbib}
\usepackage{booktabs}
\usepackage{courier}
\usepackage[compact]{titlesec}
\usepackage[capitalise]{cleveref}
\usepackage{placeins}
\usepackage[export]{adjustbox}
\usepackage{siunitx}

\newcommand{\us}[1]{^{\text{#1}}}          
\newcommand*\diff{\mathop{}\!\mathrm{d}}
\DeclareMathOperator*{\E}{\mathbb{E}}
\newcommand{\D}{\mathcal{D}}

\jname{Environmental Data Science}
\jyear{2023}

\begin{document}

\begin{Frontmatter}

\title[Article Title]{Environmental Sensor Placement with Convolutional Gaussian Neural Processes}

\author*[1]{Tom R. Andersson}\orcid{0000-0002-1556-9932}\email{tomand@bas.ac.uk}
\author[2]{Wessel P. Bruinsma}
\author[3]{Stratis Markou}
\author[4]{James Requeima}
\author[5]{Alejandro Coca-Castro}
\author[3]{Anna Vaughan}
\author[6]{Anna-Louise Ellis}
\author[7,8]{Matthew A. Lazzara}
\author[1]{Dani Jones$^\dagger$}
\author[1,5]{J. Scott Hosking$^\dagger$}
\author[2,3]{Richard E. Turner$^\dagger$}
\address[1]{\orgname{British Antarctic Survey, NERC, UKRI}}
\address*[2]{\orgname{Microsoft Research AI4Science}}
\address*[3]{\orgname{University of Cambridge}}
\address*[4]{\orgname{Vector Institute}}
\address*[5]{\orgname{The Alan Turing Institute}}
\address*[6]{\orgname{Met Office}}
\address*[7]{\orgname{University of Wisconsin-Madison}}
\address*[8]{\orgname{Madison Area Technical College}}
\address*[$^\dagger$]{\orgname{Joint senior authors}}

\received{1 February 2023}
\revised{29 March 2023}
\accepted{05 May 2023}

\authormark{Tom R. Andersson et al.}


\abstract{
Environmental sensors are crucial for monitoring weather conditions and the impacts of climate change.
However, it is challenging to place sensors in a way that maximises the informativeness of their measurements, particularly in remote regions like Antarctica.
Probabilistic machine learning models can suggest informative sensor placements by finding sites that maximally reduce prediction uncertainty.
Gaussian process (GP) models are widely used for this purpose, but they struggle with capturing complex non-stationary behaviour and scaling to large datasets. 
This paper proposes using a convolutional Gaussian neural process (ConvGNP) to address these issues.
A ConvGNP uses neural networks to parameterise a joint Gaussian distribution at arbitrary target locations, enabling flexibility and scalability.
Using simulated surface air temperature anomaly over Antarctica as training data, the ConvGNP learns spatial and seasonal non-stationarities, outperforming a non-stationary GP baseline.
In a simulated sensor placement experiment, the ConvGNP better predicts the performance boost obtained from new observations than GP baselines, leading to more informative sensor placements. 
We contrast our approach with physics-based sensor placement methods and propose future steps towards an operational sensor placement recommendation system.
Our work could help to realise environmental digital twins that actively direct measurement sampling to improve the digital representation of reality.
}

\policy{
This paper addresses the challenge of identifying intelligent sensor placements for monitoring environmental phenomena, using Antarctic air temperature anomaly as an example.
The authors propose using a recent machine learning model---a convolutional Gaussian neural process (ConvGNP)---which can capture complex non-stationary behaviour and scale to large datasets.
The ConvGNP outperforms previous data-driven approaches in simulated experiments, finding more informative and cost-effective sensor placements.
This could lead to improved decision-making for monitoring weather conditions and climate change impacts.
}

\end{Frontmatter}

\section{Introduction}
Selecting optimal locations for placing environmental sensors is an important scientific challenge.
For example, improved environmental monitoring can lead to more accurate weather forecasting \citep{weissmann_influence_2011, jung_advancing_2016}.
Further, better observation coverage can improve the representation of extreme events, climate variability, and long-term trends in reanalysis models \citep{bromwich_strong_2004} and aid their validation \citep{bracegirdle_reliability_2012}.
This is particularly important in remote regions like Antarctica, where observations are sparse \citep{jung_advancing_2016} and the cost of deploying weather stations is high \citep{lazzara_antarctic_2012},
motivating an objective model-based approach that provides an accurate notion of the informativeness of new observation locations.
This informativeness can then guide decision-making so that scientific goals are achieved with as few sensors as possible.

The above sensor placement problem has been studied extensively from a physics-based numerical modelling perspective \citep{majumdar_review_2016}.
Multiple approaches exist for estimating the value of current or new observation locations for a numerical model.
Examples include observing system simulation experiments (\citealt{hoffman_future_2016}), adjoint methods \citep{langland_estimation_2004}, and ensemble sensitivity analysis \cite[ESA;][]{torn_ensemble-based_2008}.
Using a numerical model for sensor placement comes with benefits and limitations.
One drawback is that numerical models can be biased, and this can degrade sensor placements.
This suggests that physics-based approaches could be supplemented by data-driven methods that learn statistical relationships directly from the data.

Machine learning (ML) methods also have a long history of use for experimental design and sensor placement \citep{mackay_information-based_1992, cohn_neural_1993, seo_gaussian_2000, krause_near-optimal_2008}.
First, a \textit{probabilistic model} is fit to noisy observations of an unknown function $f(\bm{x})$.
Then, \emph{active learning} is used to identify new $\bm{x}$-locations that are expected to maximally reduce the model's uncertainty about some aspect of $f(\bm{x})$.
The Gaussian process (GP; \citealt{rasmussen_gaussian_2004}) has so far been the go-to class of probabilistic model for sensor placement and the related task of Bayesian optimisation\footnote{Bayesian optimisation differs slightly from sensor placement in that the task is to find the maximum (of minimum) a black-box function $f$ rather than reduce overall uncertainty about $f$.} \citep{singh_efficient_2007, krause_near-optimal_2008, marchant_bayesian_2012, shahriari_taking_2016}.
Setting up a GP requires the user to specify a mean function (describing the expected value of the function) and a covariance function (describing how correlated the $f(\bm{x})$ values are at different $\bm{x}$-locations).
Once a GP has been initialised, conditioning it on observed data and evaluating at target locations produces a multivariate Gaussian distribution, which can be queried to search for informative sensor placements.

GPs have several compelling strengths which make them particularly amenable to small-data regimes and simple target functions.
However, modelling a climate variable with a GP is challenging due to spatiotemporal non-stationarity and large volumes of data corresponding to multiple predictor variables.
While non-stationary GP covariance functions are available (and improve sensor placement in \citealt{krause_near-optimal_2008} and \citealt{singh_modeling_2010}), this still comes with the task of choosing the right functional form and introduces a risk of overfitting \citep{fortuin_meta-learning_2020}.
Further, conditioning GPs on supplementary predictor variables (such as satellite data) is non-trivial and their computational cost scales cubically with dataset size, which becomes prohibitive with large environmental datasets.
Approximations allow GPs to scale to large data \citep{titsias_variational_2009, hensman_gaussian_2013}, but these also harm prediction quality.
The above model misspecifications can lead to uninformative or degraded sensor placements, motivating a new approach which can more faithfully capture the behaviour of complex environmental data.

Convolutional neural processes (ConvNPs) are a recent class of ML models that have shown promise in modelling environmental variables.
For example, ConvNPs can outperform a large ensemble of climate downscaling approaches \citep{vaughan_convolutional_2021, markou_practical_2022} and integrate data of gridded and point-based modalities \citep{bruinsma_autoregressive_2023}.
One variant, the convolutional Gaussian neural process \cite[ConvGNP;][]{markou_practical_2022, bruinsma2021gaussian}, uses neural networks to parameterise a joint Gaussian distribution at target locations, allowing them to scale linearly with dataset size while learning mean and covariance functions directly from the data.

In this paper, simulated atmospheric data is used to assess the ability of the ConvGNP to model a complex environmental variable and find informative sensor placements.
The paper is laid out as follows.
\Cref{section.methods} introduces the data and describes the ConvGNP model.
\Cref{section.results} compares the ConvGNP with GP baselines with three experiments: predicting unseen data, predicting the benefit of new observations, and a sensor placement toy experiment.
It is then shown how placement informativeness can be traded-off with cost using multi-objective optimisation to enable a human-in-the-loop decision-support tool.
\Cref{section.discussion} discusses limitations and possible extensions to our approach, contrasting ML-based and physics-based sensor placements.
Concluding remarks are provided in \Cref{section.conclusion}.

\section{Methods}\label{section.methods}

In this section we define the goal and data, formalise the problem tackled, and introduce the ConvGNP.

\subsection{Goal and source data}
\begin{figure}[htbp]
    \vspace{-6mm}
    \centering
    \includegraphics[trim=0cm 0.25cm 0cm 0.4cm,clip,scale=1]{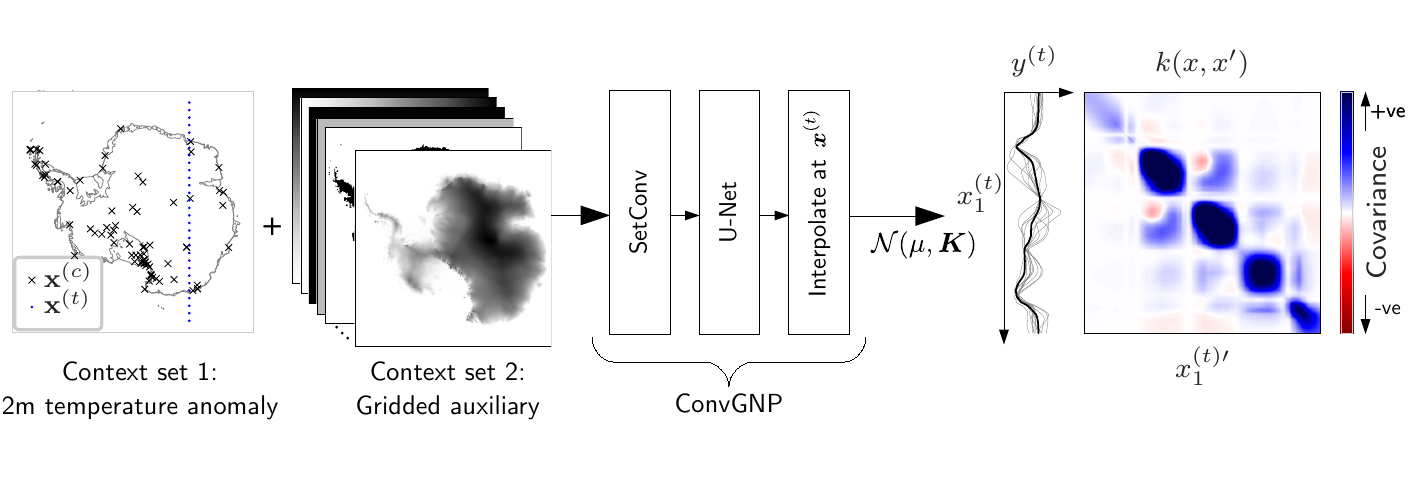}
    \vspace{-8mm}
    \caption{\textbf{The ConvGNP.} We have two context sets: ERA5 temperature anomaly observations and 6 gridded auxiliary fields, and we wish to make probabilistic predictions for temperature anomaly over a vertical line of target points (blue dotted line in left-most panel). In the ConvGNP, a SetConv layer fuses the context sets into a single gridded encoding (\Cref{fig:apx.setconv}, \citealt{gordon_convolutional_2020}). A U-Net \citep{ronneberger_u-net_2015} takes this encoded tensor as input and outputs a gridded representation, which is interpolated at target points $\bm{X}\us{(t)}$ and used to parameterise the mean and covariance of a multivariate Gaussian distribution over $\bm{y}\us{(t)}$. The output mean vector $\bm{\mu}$ is shown as a black line, with 10 Gaussian samples overlaid in grey. The heatmap of the covariance matrix $\bm{K}$ shows the magnitude of spatial covariances, with covariance decreasing close to temperature anomaly context points.}
    \label{fig:foward_pass}
    \vspace{-3mm}
\end{figure}

We use reanalysis data to analyse sensor placement abilities.
Reanalysis data are produced by fitting a numerical climate model to observations using data assimilation \citep{gettelman_future_2022}, capturing the complex dynamics of the Earth system on a regular grid.
The simulated target variable used in this study is \SI{25}{km}-resolution ERA5 daily-averaged \SI{2}{m} temperature anomaly over Antarctica (\Cref{fig:samples}a; \citealt{hersbach_era5_2020}).
For a given day of year, temperature anomalies are computed by subtracting maps of the mean daily temperature (averaged over 1950--2013) from the absolute temperature, removing the seasonal cycle.
We train a ConvGNP and a set of GP baselines to produce probabilistic spatial interpolation predictions for ERA5 temperature anomalies, assessing performance on a range of metrics.
We then perform simulated sensor placement experiments to quantitatively compare the ConvGNP's estimates of observation informativeness with that of the GP baselines and simple heuristic placement methods. 
The locations of 79 Antarctic stations that recorded temperature on February 15th, 2009 are used as the starting point for the sensor placement experiment (black crosses in \Cref{fig:foward_pass}), simulating a realistic sensor network design scenario.
Alongside inputs of ERA5 temperature anomaly observations, we also provide the ConvGNP with a second data stream on a \SI{25}{km} grid, containing surface elevation and a land mask (obtained from the BedMachine dataset; \citealt{morlighem__mathieu_measures_2020}), as well as space/time coordinate variables.
For further details on the data sources and preprocessing see Appendix \ref{apx.data_considerations}.

\subsection{Formal problem set-up}\label{section:problem}

We now formalise the problem set-up tackled in this study.
First, we make some simplifying assumptions about the data to be modelled.
We assume that data from different time steps, $\tau$, are independent, and so we will not model temporal dependencies in the data.
Further, we only consider variables that live in a 2D input space, as opposed to variables with a third input spatial dimension (e.g.~altitude or depth).
This simplifies the 3D or 4D modelling problem into a 2D one.
Models built with these assumptions can learn correlations across 2D space, but not across time and/or height, which could be important in forecasting or oceanographic applications.

At each $\tau$, there will be particular target locations $\bm{X}_\tau\us{(t)} \in \mathbb{R}^{N_t\times2}$ where we wish to predict an environmental variable $\bm{y}_\tau\us{(t)} \in \mathbb{R}^{N_t}$ (we assume that the target variable is a 1D scalar for simplicity, but this need not be the case).
Our target may be surface temperature anomaly along a line of points over Antarctica (blue dotted line in left-most panel of \Cref{fig:foward_pass}).
We call this a \emph{target set} $T_\tau=(\bm{X}_\tau\us{(t)}, \bm{y}_\tau\us{(t)})$.
The target set predictions will be made using several data streams, containing $N$-D observations $\bm{Y}_\tau\us{(c)} \in \mathbb{R}^{N_c \times N}$ at particular locations $\bm{X}_\tau\us{(c)} \in \mathbb{R}^{N_c\times2}$.
We call these data streams \emph{context sets} $(\bm{X}_\tau\us{(c)}, \bm{Y}_\tau\us{(c)})$, and write the collection of all $N_C$ context sets as
$
    C_\tau = \{ (\bm{X}_\tau\us{(c)}, \bm{Y}_\tau\us{(c)})_i \}_{i=1}^{N_C}.
$
Context sets may lie on scattered, off-grid locations (e.g.~temperature anomaly observations at black crosses in left-most panel in \Cref{fig:foward_pass}) or on a regular grid (e.g.~elevation and other auxiliary fields in the second panel of \Cref{fig:foward_pass}). 
We call the collection of context sets and the target set a \emph{task}
$
    \D_\tau = (C_\tau, T_\tau).
$
The goal is to build a ML model that takes the context sets as input and maps to probabilistic predictions for the target values $\bm{y}_\tau\us{(t)}$ given the target locations $\bm{X}_\tau\us{(t)}$.
Following \citealt{foong_meta-learning_2020}, we refer to this model as a \emph{prediction map}, $\pi$.
Once $\pi$ is set up, a sensor placement algorithm $\mathcal{S}$ will use $\pi$ to propose $K$ new placement locations $\bm{X}^* \in \mathbb{R}^{K\times2}$ based on query locations $\bm{X}\us{(s)} \in \mathbb{R}^{S\times2}$ and a set of tasks $\{\D_{\tau_j}\}_{j=1}^{J}$.
\Cref{section.sensor_placement_toy_exp} provides details on how we implement $\mathcal{S}$ in practice.

Physics-based numerical models could be framed as hard-coded prediction maps, ingesting context sets through data assimilation schemes and using physical laws to predict targets on a regular grid over space and time.
These model outputs are deterministic by default, but applying stochastic perturbations to initial conditions or model parameters induces an intractable distribution over model outputs, $p(\bm{y}\us{(t)})$, which can be sampled from to generate an ensemble of reanalyses or forecasts.
However, current numerical models do not learn directly from data.
In contrast, ML-based prediction maps will be trained from scratch to directly output a distribution over targets based on the context data.

\subsection{ConvGNP model}\label{section.convgnp}

Most ML methods are ill-suited to the problem described in \Cref{section:problem}.
Typical deep learning approaches used in environmental applications, such as convolutional neural networks, require the data to lie on a regular grid, and thus cannot handle non-gridded data (e.g.~\citealt{andersson_seasonal_2021}; \citealt{ravuri_skilful_2021}).
Recent emerging architectures such as transformers can handle off-the-grid data in principle, but in practice have used gridded data in environmental applications (e.g.~\citealt{bi_pangu-weather_2022}).
Moreover, they also need architectural changes to make predictions at previously unseen input locations.
On the other hand, Bayesian probabilistic models based on stochastic processes (such as GPs) can ingest data at arbitrary locations, but it is difficult to integrate more than one input data stream, especially when those streams are high dimensional (e.g.~supplementary satellite data which aids the prediction task).
Neural processes \cite[NPs;][]{garnelo_conditional_2018, garnelo_neural_2018} are prediction maps that address these problems by combining the modelling flexibility and scalability of neural networks with the uncertainty quantification benefits of GPs.
The ConvGNP is a particular prediction map $\pi$ whose output distribution is a correlated (joint) Gaussian with mean $\bm{\mu}$ and covariance matrix $\bm{K}$:

\vspace{-3mm}
\begin{equation}\label{eq:gnp}
    \pi(\bm{y}\us{(t)}; C, \bm{X}\us{(t)}) = \mathcal{N} (\bm{y}\us{(t)}; \bm{\mu}(C,\bm{X}\us{(t)}), \bm{K}(C,\bm{X}\us{(t)})).
\end{equation}
\vspace{-3mm}

\noindent The ConvGNP takes in the context sets $C$ and outputs a mean and non-stationary covariance function of a GP predictive, which can be queried at arbitrary target locations (\Cref{fig:foward_pass}).
It does this by first fusing the context sets into a gridded encoding using a SetConv layer \citep{gordon_convolutional_2020}.
The SetConv encoder interpolates context observations onto an internal grid with the density of observations captured by a `density channel' for each context set (example encoding shown in \Cref{fig:apx.setconv}).
This endows the model with the ability to ingest multiple predictors of various modalities (gridded and point-based) and handle missing data (\Cref{apx.subsection.setconv}).
The gridded encoding is passed to a U-Net \citep{ronneberger_u-net_2015}, which produces a \emph{representation} of the context sets with $\bm{R} = \text{U-Net}(\text{SetConv}(C))$.
The tensor $\bm{R}$ is then spatially interpolated at each target location $\bm{x}_{i}\us{(t)}$, yielding a vector $\bm{r}_i$ and enabling the model to predict at arbitrary locations.
Finally, $\bm{r}_i$ is passed to multilayer perceptrons $f$ and $\bm{g}$, parameterising the mean and covariance respectively with $\mu_i = f(\bm{r}_i)$ and $k_{ij} = \bm{g}(\bm{r}_i)\us{T} \bm{g}(\bm{r}_j)$.
This architecture results in a mean vector $\bm{\mu}$ and covariance matrix $\bm{K}$ that are \emph{functions} of $C$ and $\bm{X}\us{(t)}$ (Equation~\ref{eq:gnp}).

Constructing the covariances via a dot product leads to a low-rank covariance matrix structure, which is exploited to reduce the computational cost of predictions from cubic to linear in the number of targets.
Furthermore, the use of a SetConv to encode the context sets results in linear scaling with the number of context points.
This out-of-the-box scalability allows the ConvGNP to process 100,000 context points and predict over 100,000 target points in less than a second on a single GPU\footnote{Our ConvGNP (with 4.16M parameters) takes \SI{0.88}{s} to process a total of 100,000 context points (21,600 temperature points and 78,400 gridded auxiliary points) and predict over 100,000 target points on a 16 GB NVIDIA A4 GPU using TensorFlow's eager mode.}.

NPs can be considered \emph{meta-learning} models \citep{foong_meta-learning_2020} which \emph{learn how to learn}, mapping directly from context sets to predictions without requiring retraining when presented with new tasks.
This is useful in environmental sciences because it enables learning statistical relationships (such as correlations) that depend on the context observations.
In contrast, conventional supervised learning models, such as GPs, instead learn fixed statistical relationships which do not depend on the context observations.

\subsection{Training the ConvGNP}

Training tasks $\mathcal{D}_\tau$ are generated by first sampling the day $\tau$ randomly from the training period, 1950--2013.
Then, ERA5 grid cells are sampled uniformly at random across the entire 280 $\times$ 280 input space, with the number of ERA5 temperature anomaly context and target points drawn uniformly at random from $N_c \in \{5, 6, \dots, 500\}$ and $N_t \in \{3000, 3001, \dots, 4000\}$.
The ConvGNP is trained to minimise the negative log-likelihood (NLL) of target values $\bm{y}\us{(t)}_\tau$ under its output Gaussian distribution using the Adam optimiser. 
After each training epoch, the model is checkpointed if an improvement is made to the mean NLL on validation tasks from 2014--2017.
For further model and training details see \Cref{apx.convgnp}.

\begin{figure}[htbp]
    \vspace{-2mm}
    \centering
    \includegraphics[scale=1]{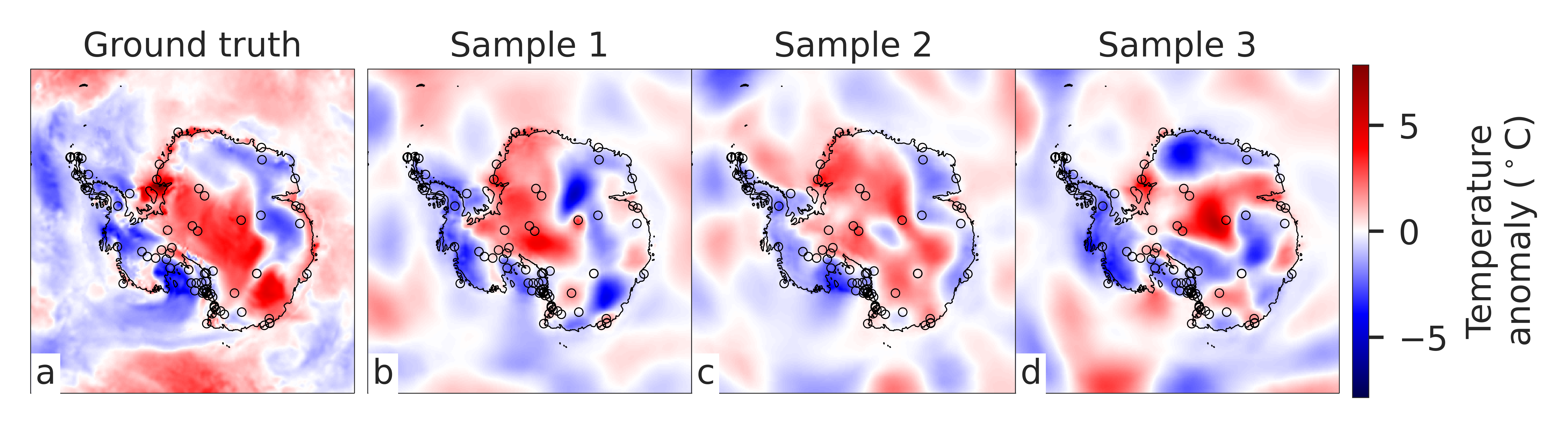}
    \vspace{-3mm}
    \caption{\textbf{The ConvGNP extrapolates plausible scenarios away from data.} \textbf{a}, ERA5 \SI{2}{m} temperature anomaly on January 1st 2018; \textbf{b-d}, ConvGNP samples with ERA5 temperature anomaly context points at Antarctic station locations (black circles). Comparing colours within the black circles across plots shows that the ConvGNP interpolates context observations.}
    \label{fig:samples}
    \vspace{-2mm}
\end{figure}

\begin{figure}[htbp]
    \vspace{-2mm}
    \centering
    \includegraphics[trim=0.6cm 0cm 0cm 0cm,clip,scale=0.9]{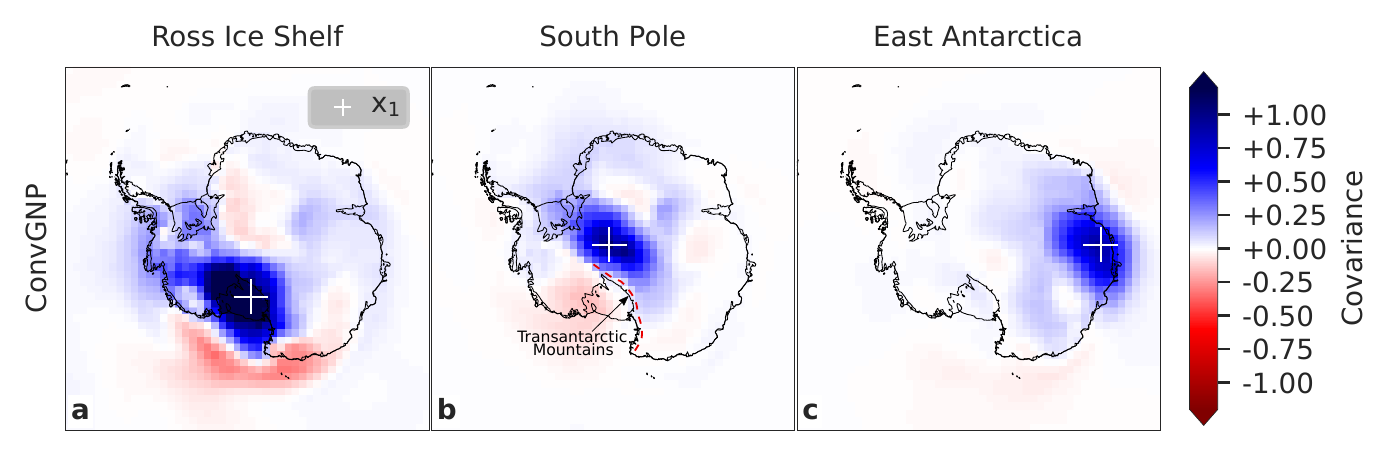}
    \vspace{-2mm}
    \caption{\textbf{The ConvGNP learns spatially-varying covariance structure.}
    Prior covariance function, $k(\bm{x}_1, \bm{x}_2)$, with $\bm{x}_1$ fixed at the white plus location and $\bm{x}_2$ varying over the grid. Plots are shown for three different $\bm{x}_1$-locations (the Ross Ice Shelf, the South Pole, and East Antarctica) for the 1st of June. The most prominent section of the Transantarctic Mountains is indicated by the red dashed line in \textbf{b}.}
    \label{fig:cov}
    \vspace{-2mm}
\end{figure}

Once trained in this manner, the ConvGNP outputs expressive, non-stationary mean and covariance functions.
When conditioning the ConvGNP on ERA5 temperature anomaly observations and drawing Gaussian samples on a regular grid, the samples interpolate observations at the context points and extrapolate plausible scenarios away from them (\Cref{fig:samples}).
Running the ConvGNP on a regular grid with no temperature anomaly observations reveals the prior covariance structure learned by the model (\Cref{fig:cov}).
The ConvGNP leverages the gridded auxiliary fields and day of year inputs from the second context set to output highly non-stationary spatial dependencies in surface temperature, such as sharp drops in covariance over the coastline (\Cref{fig:cov}a-c), anticorrelation (\Cref{fig:cov}a), and decorrelation over the Transantarctic Mountains (\Cref{fig:cov}b).
In Appendix \ref{apx.section.nonstationarity} we contrast this with GP prior covariances and further show that the ConvGNP learns seasonally-varying spatial correlation (\Cref{fig:apx.cov}--\Cref{fig:apx.corr_diff}).

\section{Results}\label{section.results}

We evaluate the ConvGNP's ability to model ERA5 \SI{2}{m} daily-average surface temperature anomaly through a range of experiments, using GP baselines with both non-stationary and stationary covariance functions.
We use three GP baselines with different non-isotropic covariance functions: the exponentiated quadratic (EQ), the rational quadratic (RQ), and the Gibbs kernel.
The EQ and RQ are stationary because the covariance depends only on the difference between two input points, $k(\bm{x}, \bm{x}') = k(\bm{x} - \bm{x}')$.
The Gibbs covariance function is a more sophisticated, non-stationary baseline, where the correlation length scale is allowed to vary over space (\Cref{fig:apx.gibbs_l_x}).
As noted in \Cref{section.convgnp}, there is no simple way to condition vanilla GP models on multiple context sets; the GP baselines can only ingest the context set containing the ERA5 observations and not the second, auxiliary context set.
For more details on the GPs, including their covariance functions and training procedure, see \Cref{apx.section.benchmarks}.

\subsection{Performance on unseen data}\label{section:test_results}

To assess the models' abilities to predict unseen data, 30,618 tasks are generated from unseen test years 2018--2019 by sampling ERA5 grid cells uniformly at random with number of targets $N_t=2,000$ and a range of context set sizes $N_c \in \{0, 25, 50, \dots, 500\}$ (Appendix \ref{apx.section.dtau_generation}).
For each task, we compute three performance metrics of increasing complexity.
The first metric, the root mean squared error (RMSE), simply measures the difference between the model's mean prediction and the true values.
The second metric, the mean marginal NLL, includes the variances of the model's point-wise Gaussian distributions, measuring how confident and well-calibrated the marginal distributions are.
The third metric, the joint NLL, uses the model's full joint Gaussian distribution, measuring how likely the true $\bm{y}\us{(t)}$ vector is under the model.
This quantifies the reliability of the model's off-diagonal spatial correlations as well as its marginal variances.

\begin{figure}[htbp]
    \vspace{-2mm}
    \centering
    \includegraphics[scale=0.95]{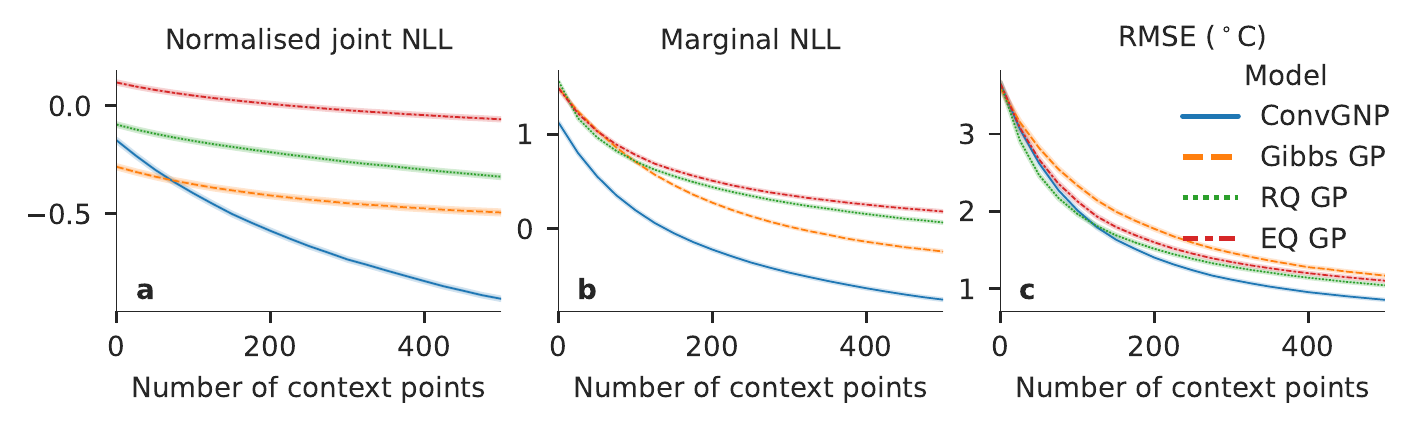}
    \vspace{-3mm}
    \caption{\textbf{Test set results.} Mean metric values versus number of context points on the test set. The joint negative log-likelihood (NLL) is normalised by the number of targets. Error bars are standard errors.}
    \label{fig:results}
    \vspace{-4mm}
\end{figure}

In general, the ConvGNP performs best, followed by the Gibbs GP, the RQ GP, and finally the EQ GP (\Cref{fig:results}).
There are some exceptions to this trend.
For example, the models produce similar RMSEs for $N_c<100$.
This is likely because for small $N_c$ the models revert to zero-mean predictions away from context points (matching the zero-mean of temperature anomaly over the training period).
Another exception is that the Gibbs GP outperforms the ConvGNP on joint NLL for small $N_c$.
This may be because the ConvGNP's training process biases learning towards `easier' tasks (where $N_c$ is larger).
Alternatively, the ConvGNP's low-rank covariance parameterisation could be poorly suited to small $N_c$.
However, with increasing $N_c$, the ConvGNP significantly outperforms all three GP baselines across all three metrics, with its performance improving at a faster rate with added data.
When averaging the results across $N_c$, the ConvGNP significantly outperforms all three GP baselines for each metric (\Cref{tab:apx.antarctica-results}).
We further find that the ConvGNP's marginal distributions are substantially sharper and better calibrated than the GP baselines (\Cref{fig:apx.marginal_calib} and \Cref{fig:apx.marginal_sharpness}), which is an important goal for probabilistic models \citep{gneiting_probabilistic_2007}.
Well-calibrated uncertainties are also key for active learning, which is explored below in \Cref{section.sensor_placement_toy_exp}.

\subsection{Sensor placement}\label{section.sensor_placement_toy_exp}

Following previous works \citep{krause_near-optimal_2008}, we pose sensor placement as a discrete optimisation problem.
The task is to propose a subset of $K$ sensor placement locations, $\bm{X}^*$, from a set of $S$ search locations, $\bm{X}\us{(s)}$.
In practice, to avoid the infeasible combinatorial cost of searching over multiple placements jointly, a \emph{greedy} approximation is made by selecting one sensor placement at a time.
Within a greedy iteration, a value is assigned to each query location $\bm{x}\us{(s)}_i$ using an \emph{acquisition function}, $\alpha(\bm{x}\us{(s)}_i, \tau)$, specifying the utility of a new observation at $\bm{x}\us{(s)}_i$ for time $\tau$, which we average over $J$ dates: 
\vspace{-3mm}
\begin{equation}\label{eq:alpha_avg}
    \vspace{-2mm}
    \alpha(\bm{x}\us{(s)}_i) = \frac{1}{J}\sum_{j=1}^J \alpha(\bm{x}\us{(s)}_i, \tau_j).
\end{equation}
\vspace{-3mm}

\noindent We use five acquisition functions which are to be maximised, defining a set of placement criteria (mathematical definitions are provided in \Cref{apx.section.acquisition_functions}):

\textbf{\texttt{JointMI}}: mutual information (MI) between the model's prediction and the query sensor observation, imputing the missing value with the model's mean at the query location, $\bar{y}_{\tau, i}\us{(s)}$.\footnote{A better approach would be to draw Monte Carlo samples over $y_{\tau, i}\us{(s)}$, although this would be more costly -- see \Cref{apx.section.acquisition_functions}.}
This criterion attempts to minimise the model's joint entropy by minimising the log-determinant of the output covariance matrix, balancing minimising marginal variances with maximising correlation magnitude, which can be viewed as minimising uncertainty about the spatial patterns \citep{mackay_information-based_1992}.
The joint MI has been used frequently in past work \citep{lindley_measure_1956, krause_near-optimal_2008, schmidt_sequential_2019}.

\textbf{\texttt{MarginalMI}}: as above, but ignoring the off-diagonal elements in the models' Gaussian distributions and considering only the diagonal (marginal) entries.
This criterion attempts to minimise the model's marginal entropy by minimising the log-variances in the output distribution.

\textbf{\texttt{DeltaVar}}: decrease in average marginal variance in the output distribution (similar to \texttt{MarginalMI} but using absolute variances rather than log-variances).
Previous works have used this criterion for active learning both with neural networks \citep{cohn_neural_1993} and GPs \citep{seo_gaussian_2000}.

\textbf{\texttt{ContextDist}}: distance to the closest sensor.
This is a simple heuristic which proposes placements as far away as possible from the current observations.
While this is a strong baseline, non-stationarities in the data will mean that it is sub-optimal.
For example, a high density of sensors will be needed in areas where correlation length scales are short, and a low density where they are large.
Therefore, the optimal sensor placement strategy should differ from and outperform this approach.

\textbf{\texttt{Random}}: uniform white noise function (i.e.~placing sensors randomly).
The performance of this criterion reflects the average benefit of adding new observations for a given model and context set.

\vspace{1mm}

\begin{figure}[htbp]
    \vspace{-2mm}
    \centering
    \includegraphics[trim=0cm 0cm 0cm 0.7cm,clip,scale=0.97]{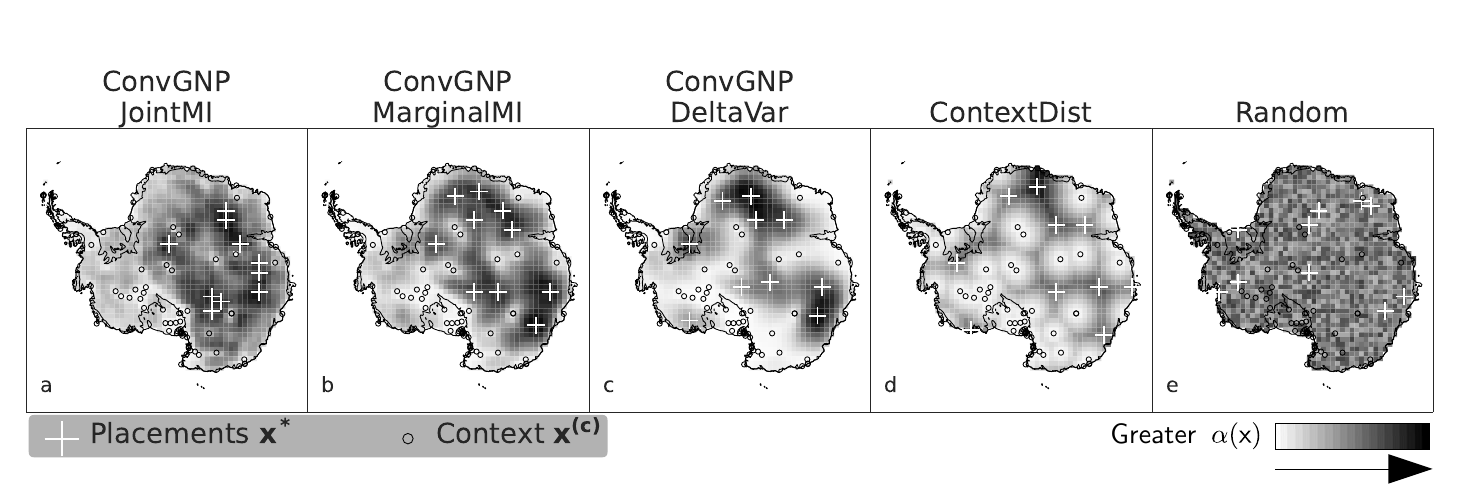}
    \vspace{-5mm}
    \caption{\textbf{Acquisition functions and sensor placements for the ConvGNP and heuristic baselines.} Maps of acquisition function values $\alpha(\bm{x}\us{(s)}_i)$ for the initial $k=1$ greedy iteration. The initial context set $\bm{X}^{(c)}$ is derived from real Antarctic station locations (black circles). Running the sensor placement algorithm for $K=10$ sensor placements results in the proposed sensor placements $\bm{X}^*$ (white pluses). Each pixel is $100\times$\SI{100}{km}.}
    \label{fig:sp}
    \vspace{-4mm}
\end{figure}

The target locations $\bm{X}_\tau\us{(t)}$ and search locations $\bm{X}\us{(s)}$ are both defined on a \SI{100}{km} grid over Antarctica, resulting in $N_t=S=1,365$ targets and possible placement locations.
The context set locations $\bm{X}_\tau\us{(c)}$ are fixed at Antarctic temperature station locations (black circles in \Cref{fig:sp}) to simulate a realistic network design scenario.
We use $J=105$ dates from the validation period (2014--2017), sampled at a 14-day interval, to compute the acquisition functions.
Heatmaps showing the above five acquisition functions on the $\bm{X}\us{(s)}$ grid, using the ConvGNP for the model underlying the three uncertainty-reduction acquisition functions, are shown in \Cref{fig:sp}.
There are interesting differences between the model-based acquisition functions of the ConvGNP, the Gibbs GP, and the EQ GP (\Cref{fig:apx.sp_all}).

\subsubsection{Oracle acquisition function experiment}\label{section.oracle_acquisition_functions}

By using sensor placement criteria that reduce uncertainty in the model's predictions, one hopes that predictions also become more accurate in some way.
For example, the entropy of the model's predictive distribution is the expected NLL of the data under the model, so the decrease in entropy from a new observation (i.e.~the MI) should relate to the NLL improvement -- assuming the model is well-specified for the data.
Further, if the model's marginal distributions are well-calibrated, marginal variance relates to expected squared error.
Therefore, the \texttt{JointMI}, \texttt{MarginalMI}, and \texttt{DeltaVar} acquisition functions should relate to improvements in joint NLL, marginal NLL, and RMSE, respectively. 
However, in general, the strength of these relationships are unknown.
In the toy setting of this study, where ERA5 is treated as ground truth and is known everywhere, these relationships can be examined empirically.

\begin{figure}[htbp]
    \vspace{-0.25cm}
    \centering
    \includegraphics[scale=0.85]{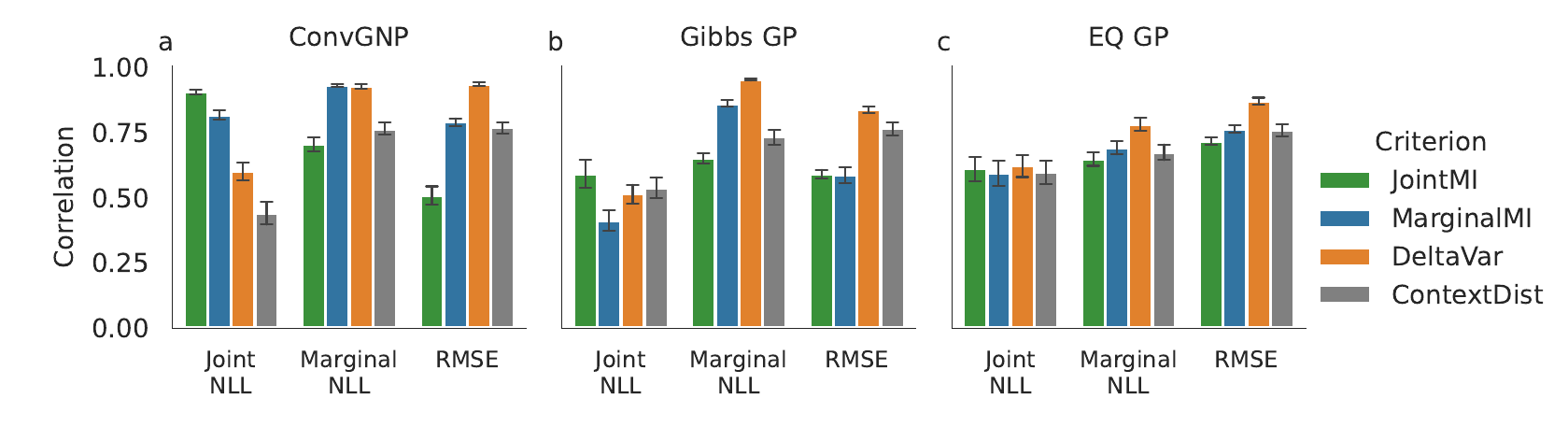}
    \vspace{-10mm}
    \caption{\textbf{The ConvGNP reliably predicts the benefit of new observations}. Correlation between model-based and oracle acquisition functions, $\bm{\alpha}(\bm{X}\us{(s)})$ and $\bm{\alpha_}\text{oracle}(\bm{X}\us{(s)})$. Error bars indicate the 2.5\%--97.5\% quantiles from 5000 bootstrapped correlation values, computed by resampling the 1365 pairs of points with replacement, measuring how spatially consistent the correlation is across the search space $\bm{X}\us{(s)}$   .}
    \label{fig:oracle_correlation_results}
    \vspace{-2mm}
\end{figure}

We compare the ability of the ConvGNP, Gibbs GP, and EQ GP to predict the benefit of new observations based on the \texttt{JointMI}, \texttt{MarginalMI}, and \texttt{DeltaVar} acquisition functions, using \texttt{ContextDist} as a na\"ive baseline.
The true benefit of observations is determined using \emph{oracle} acquisition functions, $\alpha_\text{oracle}$, where the true ERA5 value is revealed at $\bm{x}\us{(s)}_i$ and the average performance gain on the target set is measured for each metric: joint NLL, marginal NLL, and RMSE (\Cref{apx.section.oracle_acquisition_functions}).
Computing non-oracle and oracle acquisition functions at all $S$ query locations produces vectors, $\bm{\alpha}(\bm{X}\us{(s)})$ and $\bm{\alpha}_\text{oracle}(\bm{X}\us{(s)})$.
The Pearson correlation $r=\text{corr}(\bm{\alpha}(\bm{X}\us{(s)}), \bm{\alpha_}\text{oracle}(\bm{X}\us{(s)}))$ between these vectors quantifies how strong the relationship is for a given model, acquisition function, and metric. 
With the context set initialised at Antarctic station locations (\Cref{section.sensor_placement_toy_exp}), the ConvGNP's joint MI achieves the best correlation with its joint NLL improvement ($r=0.90$), as for its marginal MI with its marginal NLL improvement ($r=0.93$) and its change in variance with its RMSE improvement ($r=0.93$) (\Cref{fig:oracle_correlation_results}a), substantially outperforming the \texttt{ContextDist} baseline in each case.
The Gibbs GP's acquisition functions are less robust at predicting performance gain, with the joint MI being particularly poor at predicting joint NLL improvement (\Cref{fig:oracle_correlation_results}b).
The EQ GP's model-based acquisition functions all perform similarly to \texttt{ContextDist} for each metric (\Cref{fig:oracle_correlation_results}c), which is likely an artifact of its stationary covariance function.

We repeat the above analysis using the Kendall rank correlation coefficient, $\kappa$, which measures the similarity between the rankings of $\bm{\alpha}$ and $\bm{\alpha}_\text{oracle}$ by computing the fraction of all pairs of search points $(\bm{x}\us{(s)}_i, \bm{x}\us{(s)}_j)$ that are ordered the same way in the two rankings and normalising this fraction to lie in $(-1, 1)$ (Equation~\ref{eq:apx.kendall2}).
The findings are very similar to the Pearson correlation results above: only the ConvGNP has good alignment between acquisition functions and metrics, with \texttt{JointMI}, \texttt{MarginalMI}, and \texttt{DeltaVar} obtaining the best $\kappa$-values for joint NLL ($\kappa=0.74$), marginal NLL ($\kappa=0.82$), and RMSE ($\kappa=0.84$), respectively (\Cref{fig:apx.oracle_kendalltau_results}). 

These results indicate that the ConvGNP can robustly predict performance gain, unlike the GP baselines.
\Cref{apx.section.oracle_placement_results} provides more detailed plots from this experiment, including the acquisition function heatmaps (\Cref{fig:apx.oracle_convgnp}--\Cref{fig:apx.oracle_eq}) and scatter plots for all the oracle/non-oracle pairs underlying \Cref{fig:oracle_correlation_results} (\Cref{fig:apx.oracle_corr_nll}--\Cref{fig:apx.oracle_corr_rmse}).

\subsubsection{Sensor placement experiment}\label{section.sensor_placement}

We now run a simulated greedy sensor placement experiment.
After $\alpha(\bm{x}\us{(s)}_i)$ is computed for $i=(1, \dots, S)$, the $i^*$ corresponding to the maximum value is selected.
The corresponding input $x_{i^*}\us{(s)}$ is then appended to its context set, $\bm{X}_\tau\us{(c)} \to \{ \bm{X}_\tau\us{(c)}, x_{i^*}\us{(s)} \}$.
If $\alpha$ depends on the context $y$-values, we fill the missing observation with the model mean, $\bm{y}_\tau\us{(c)} \to \{ \bm{y}_\tau\us{(c)}, \bar{y}_{\tau, i^*}\us{(s)} \}$, where $\bar{y}_{\tau, i}\us{(s)}$ is the model's mean at $\bm{x}\us{(s)}_i$ for time $\tau$.
This process is repeated until $K=10$ placements have been made.
To evaluate placement quality, we reveal ERA5 values to the models at the proposed sites and compute performance metrics over test dates 2018--2019 with a \SI{100}{km} target grid. 
See \Cref{apx.section.placement} for full experiment details.

\begin{figure}[htbp]
    \vspace{-2mm}
    \centering
    \includegraphics[scale=0.9]{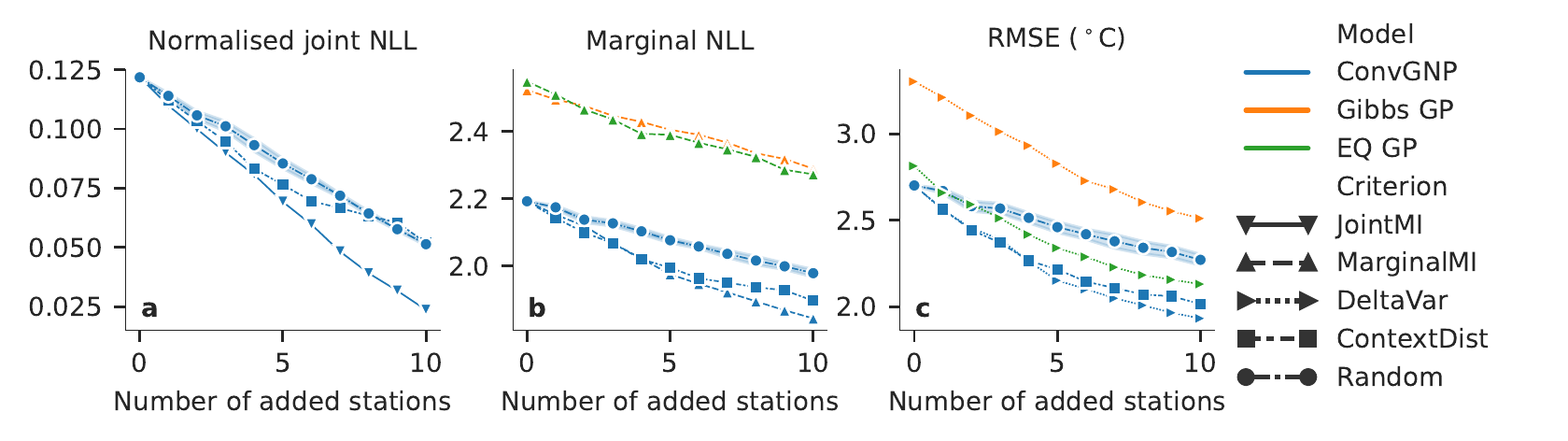}
    \vspace{-3mm}
    \caption{\textbf{Sensor placement results.} Performance metrics on the sensor placement test data versus number of stations revealed to the models. Results are averaged over 243 dates in 2018--2019, with targets defined on a \SI{100}{km} grid over Antarctica. For simplicity, we only plot the model-based criterion that targets the plotted metric. The GP baselines are shown on the marginal negative log-likelihood (NLL) and RMSE panels. For the joint NLL, the GP baselines perform far worse than the ConvGNP and are not shown. The confidence interval of \texttt{Random} is the standard error from 5 random placements.} 
    \label{fig:sp_results}
    \vspace{-5mm}
\end{figure}

The ConvGNP's \texttt{JointMI}, \texttt{MarginalMI}, and \texttt{DeltaVar} placements substantially outperform \texttt{ContextDist} for the metrics they target by the 5th placement onwards (\Cref{fig:sp_results}a--c; \Cref{fig:apx.sp_results}a,d,g), and lead to greater performance improvements by the $K=10$th placement than both of the GP baselines (\Cref{fig:apx.sp_results_sharey}).\footnote{The only exception to this is the Gibbs GP's \texttt{DeltaVar}, which improves its RMSE by \SI{0.79}{\celsius} compared with \SI{0.77}{\celsius} for the ConvGNP.
However, the Gibbs GP starts off with an RMSE that is \SI{0.60}{\celsius} worse than the ConvGNP (\Cref{fig:apx.sp_results_sharey}).}
This is despite the ConvGNP starting off with better performance than both of the GP baselines for each metric.
Furthermore, the proposed locations from the ConvGNP model-based criteria differ greatly from \texttt{ContextDist} (\Cref{fig:sp}a--d), with the \texttt{JointMI} placements being notably clustered together (\Cref{fig:sp}a).
In contrast, the EQ GP's model-based criteria propose diffuse placements (\Cref{fig:apx.sp_all}g--i) which are strikingly similar to those of \texttt{ContextDist}.
With the EQ GP's na\"ive stationary covariance, minimising uncertainty simply maximises distance from current observations, which is not a cost-effective placement strategy.
Future work should repeat these experiments with different initial sensor network configurations to assess the robustness of these results.

\subsubsection{Multi-objective optimisation for finding cost-effective sensor placements}
\begin{figure}[htbp]
    \vspace{-4mm}
    \centering
    \includegraphics[trim=0cm 0cm 0cm 0.5cm,clip,scale=0.9]{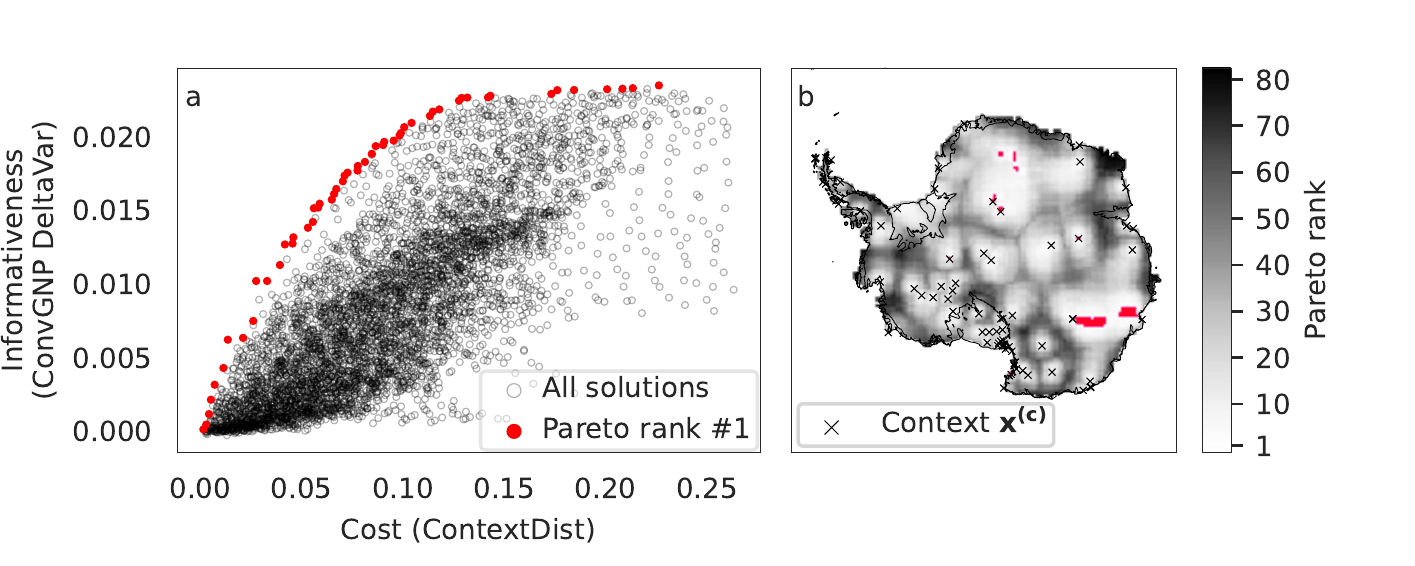}
    \vspace{-4mm}
    \caption{\textbf{Trading off informativeness with cost}. Accounting for sensor placement cost using multi-objective Pareto optimisation, maximising the ConvGNP's \texttt{DeltaVar} (a proxy for informativeness) and minimising \texttt{ContextDist} (a proxy for cost). \textbf{a}, Scatter plot showing all pairs of informativeness and cost values. \textbf{b}, Heatmap of Pareto rank. The rank-1 Pareto set is highlighted in red for both plots.}
    \label{fig:pareto}
\end{figure}

In practice, the scientific goals of sensor placement must be reconciled with cost and safety considerations, which are key concerns in Antarctic fieldwork \citep{lazzara_antarctic_2012} and will likely override the model's optimal siting recommendations $\bm{X}^*$.
In this case, it is crucial that the model can faithfully predict observation informativeness across the entire search space $\bm{X}\us{(s)}$, not just at the optimal sites $\bm{X}^*$, so that informativeness can be traded-off with cost.
Leveraging our findings from \Cref{section.oracle_acquisition_functions} that the ConvGNP's \texttt{DeltaVar} is a robust indicator of RMSE and marginal NLL improvement (\Cref{fig:oracle_correlation_results}a), we demonstrate a toy example of multi-objective optimisation with \texttt{DeltaVar} as a proxy for informativeness and \texttt{ContextDist} as a proxy for cost.
One way of integrating cost in the optimisation is to constrain the search such that the total cost is within a pre-defined budget \citep{sviridenko_note_2004, krause_near-optimal_2006}.
Alternatively, cost can be traded off with informativeness in the objective, allowing for unconstrained optimisation.
We use Pareto optimisation for this purpose, which identifies a set of `Pareto optimal' sites corresponding to points where the informativeness cannot be improved without an increase to the cost.
These rank-1 points can then be removed, the Pareto optimal set computed again, and so on until all sites have been assigned a Pareto rank (\Cref{fig:pareto}).
This procedure trivially generalises to multiple objectives and could underlie a future operational, human-in-the-loop sensor placement recommendation system that leverages an accurate cost model to guide expert decision-making. 

\section{Discussion}\label{section.discussion}

In this study, we trained a ConvGNP regression model to spatially interpolate ERA5 Antarctic \SI{2}{m} temperature anomaly.
The ConvGNP learned seasonally-varying non-stationary spatial covariance by leveraging a second data stream (`context set' in meta-learning language) containing auxiliary predictor variables, such as the day of year and surface elevation. 
The more flexible architecture and second data stream allow the ConvGNP to make substantially better probabilistic predictions on test data than those of GP baselines, including a GP with a non-stationary covariance function.
A simulated sensor placement experiment was devised with context ERA5 observations initialised at real Antarctic station locations.
New sensor placements were evaluated via the reduction in model prediction uncertainty over the Antarctic continent, with different measures of uncertainty targeting different performance metrics.
For each of these uncertainty-based acquisition functions, the ConvGNP predicts its true performance metric gain from new observations substantially more accurately than GP baselines.
This leads to informative new sensor placements that improve the ConvGNP's performance metrics on test data by a wider margin than the GP baselines, despite the ConvGNP starting off with more performant predictions and thus having less room for improvement.
These findings are notable given that GPs have a long history of use in geostatistics under the term `kriging' \citep{cressie_statistics_1993} and are frequently used for sensor placement.
Equipped with a robust measure of placement informativeness from the ConvGNP, multi-objective Pareto optimisation could be used to account for sensor placement cost, pruning a large search space of possible locations into a smaller set of cost-effective sites which can be considered by human experts.
Our approach can readily be applied to other geographies and climate variables by fitting a ConvGNP to existing reanalysis data and running a greedy sensor placement algorithm, such as the ones outlined in this work.
However, there are some limitations to this approach, which we highlight below alongside recommendations for future work.

\subsection{Limitations}

\paragraph{Not accounting for real-world observations.}
The main limitation of the current approach is that by training the ConvGNP to spatially interpolate noise-free reanalysis output instead of real-world observations, the model measures the informativeness of reanalysis data and not of real-world observations.
Two consequences arise from this shortcoming.
First, the model does not account for real-world sensor noise.
A simple way to alleviate this issue would be to simulate sensor noise by training the ConvGNP with varying levels of i.i.d.~Gaussian noise added to the ERA5 context points, which could be explored in future work.
The second consequence is that bias and coarse spatial resolution in the reanalysis data are reflected in the ConvGNP's predictions.
One way to deal with this would be to train with observational data.
However, real \textit{in-situ} environmental sensor observations can be sparse in space or time, which brings a risk of spatial overfitting when used as training data for highly flexible models like the ConvGNP.
An interesting potential solution is to pre-train the ConvGNP on simulated data and fine-tune it on observational data.
Provided sufficient observational data for training, the fine-tuning phase would correct some of the simulator biases and lead to a better representation of the target variable.

\vspace{-1mm}
\paragraph{The ConvGNP must learn how to condition on observations.}
The ConvGNP is directly trained to output a GP predictive, which is different from specifying a GP prior and then conditioning that prior on context observations using Bayes' rule.
The ConvGNP's neural networks can learn non-Bayesian conditioning mechanics from the data, which brings greater flexibility at the cost of increased training requirements.
Nevertheless, provided sufficient training data and an appropriate training scheme, the ConvGNP's conditioning flexibility is better-suited to complex environmental data than similar approaches like deep kernel learning (\citealt{wilson2015deep}), where neural networks learn non-stationary prior GP covariance functions from data and then use standard Bayes' rule conditioning to compute posterior predictives (\Cref{apx.section.dkm}).
However, if insufficient data is available to train a flexible model like the ConvGNP, a more appropriate choice would be a less data-hungry model with stronger inductive biases and better-quantified epistemic uncertainty, such as a latent GP \citep{patel_uncertainty_2022}.

\subsection{Future work}

\paragraph{Possible extensions}
Going forwards, there are several possible extensions to this work with simple modifications to our approach.
For example, the ConvGNP can be used to rank the value of current stations \citep{tardif_assessing_2022}, which could identify redundant stations that can be moved to more valuable locations.
Alternatively, the model could be set up in \textit{forecasting} mode, with the target data being some number of discrete time steps ahead of the context data.
The same greedy sensor placement algorithms can then be used to find station sites that minimise forecast uncertainty, which is important for supporting safety-critical operations in remote regions like Antarctica that depend on reliable weather forecasts \citep{lazzara_antarctic_2012, hakim_optimal_2020}.
Another exciting avenue is to build a ConvGNP that can propose optimal trajectories for a fleet of moving robots (e.g.~autonomous underwater vehicles) \citep{singh_efficient_2007, marchant_bayesian_2012}.
One way to do this is to have two context sets of the target variable: one for the current time step ($\tau=0$) and another for the next time step ($\tau=+1$).
This model can propose perturbations to the robot locations from $\tau=0$ to $\tau=+1$ (within speed limits) that minimise prediction uncertainty at $\tau=+1$.
Trajectories can then be formed by running this model autoregressively. 
To extend our approach to non-Gaussian variables, our analysis could be repeated with models that output non-Gaussian distributions, such as convolutional latent neural processes \citep{foong_meta-learning_2020}, normalising flows \citep{durkan_neural_2019}, or autoregressive ConvCNPs \citep{bruinsma_autoregressive_2023}.  
In general, future work should explore training and architecture schemes that enable learning from multiple heterogeneous data sources, such as simulated data, satellite observations, and \textit{in situ} stations.
Foundation modelling approaches have recently shown substantial promise in this area \citep{nguyen_climax_2023} and could be explored with ConvNPs.

\paragraph{Comparison and integration with physics-based sensor placement methods}
As with any model-based sensor placement approach, the ConvGNP's measure of informativeness depends on the model itself.
In general, an observation with high impact on the uncertainty of one model may have little impact on the uncertainty of another model.
This raises interesting questions about which model should be trusted, particularly for models based on very different principles such as data-driven and numerical models.
It would be insightful to examine the level of agreement or disagreement between the informativeness estimates of ML and physics-based models.
Agreement would suggest that the informativeness predicted by the causal dynamics of the numerical model is also statistically evident in the training data of the ML model.
However, a blocker to such intercomparison studies is the minimal overlap between the physics-based and ML sensor placement literatures. 
Future work should trace explicit links between these distinct research worlds to translate differing terminologies and facilitate the cross-pollination of ideas.
For example, we identified potential ML analogues for several physics-based observing system design approaches: ablation-based variable importance methods \citep{fisher_all_2019} for  observing system experiments \cite[OSEs;][]{boullot_observation_2016}; gradient-based saliency methods \citep{bach_pixel-wise_2015} for adjoint modelling \citep{langland_estimation_2004}; and uncertainty-based active learning \citep{krause_near-optimal_2008} for ESA \citep{torn_ensemble-based_2008}.
Here we remark only on the latter, where we note a striking similarity.
In ESA, sensor placement informativeness is measured by assimilating query observations into a numerical model and computing the reduction in ensemble member variance for a target quantity.
This approach has been used for Antarctic temperature sensor placement in previous studies \citep{hakim_optimal_2020, tardif_assessing_2022} with a goal of minimising the total marginal variance of Antarctic surface temperature in ensemble member samples from a numerical model, which can be seen as a Monte Carlo estimate of the \texttt{DeltaVar} acquisition function used in this study.
This similarity makes ESA a ripe starting point for comparing the sensor informativeness estimates of numerical and ML models in future work.
Other than simply comparing ML-based and physics-based sensor placement methods, future work could also integrate the two.
For example, although the ConvGNP lacks the causal grounding of dynamical models, it can run orders of magnitude faster.
Thus, the ConvGNP could nominate a few observation locations from a large search space to be analysed by more expensive physics-based techniques such as adjoint sensitivity \citep{loose_quantifying_2020,loose_leveraging_2021}.

\newpage
\section{Conclusion}\label{section.conclusion}

In current numerical weather prediction and reanalysis systems, observations improve models but not vice versa \citep{gettelman_future_2022}.
Recent calls for environmental `digital twins' have highlighted the potential to improve model predictions by using the model to actively drive data capture, thus coupling the physical world with the digital twin \citep{blair_digital_2021, gettelman_future_2022}.
This coupling could be achieved through active learning with scalable and flexible ML models. 
The ConvGNP is one such model, with a range of capabilities that aid modelling complex spatiotemporal climate variables.
These include an ability to ingest multiple predictors of various modalities (gridded and off-grid) and learn arbitrary mean and covariance functions from raw data.
This study found that the ConvGNP can robustly evaluate the informativeness of new observation sites, unlike GP baselines, using simulated Antarctic air temperature anomaly as a proof-of-concept.
By providing a faithful notion of observation informativeness, the ConvGNP could underlie an operational, human-in-the-loop sensor placement recommendation tool which can find cost-effective locations for new measurements that substantially reduce model uncertainty and increase model accuracy. 
We see our approach as complementary to existing physics-based methods, with interesting avenues for comparison and integration in future.

\begin{Backmatter}

\paragraph{Acknowledgements}
We thank Markus Kaiser, Marta Garnelo, Samantha Adams, Kevin Murphy, Anton Geraschenko, Michael Brenner, and Elre Oldewage for insightful early discussions and feedback on this work.
We also thank Steve Colwell for assistance with accessing the Antarctic station data.
Finally, we thank the two anonymous Climate Informatics 2023 reviewers and the anonymous reviewer from the NeurIPS 2022 Workshop on Gaussian Processes, Spatiotemporal Modeling, and Decision-making Systems for suggestions that improved this manuscript.

\paragraph{Funding Statement}

This work was supported by Wave 1 of The UKRI Strategic Priorities Fund under the EPSRC Grant EP/W006022/1, particularly the AI for Science theme within that grant \& The Alan Turing Institute.
This research was conducted while JR and WPB were students at the University of Cambridge, where WPB was supported by the Engineering and Physical Sciences Research Council (studentship number 10436152).
RET is supported by Google, Amazon, ARM, Improbable and EPSRC grant EP/T005386/1.
DJ is supported by a UKRI Future Leaders Fellowship (MR/T020822/1).
MAL is supported via the US National Science Foundation, grant number 1924730.

\paragraph{Competing Interests}
None

\paragraph{Code and Data Availability Statement}
The code to reproduce this paper's results will be released at \href{https://github.com/tom-andersson/EDS2022-convgnp-sensor-placement}{https://github.com/tom-andersson/EDS2022-convgnp-sensor-placement}.
We are also currently developing a generic Python package for modelling environmental data with neural processes, which we aim to release in future at \href{https://github.com/tom-andersson/deepsensor}{https://github.com/tom-andersson/deepsensor}.
The ConvGNP was implemented using the Python package \texttt{neuralprocesses} (\href{https://github.com/wesselb/neuralprocesses}{https://github.com/wesselb/neuralprocesses}).
All GPs were implemented using the Python package \texttt{stheno} (\href{https://github.com/wesselb/stheno}{https://github.com/wesselb/stheno}) and optimised using the Python package \texttt{varz} (\href{https://github.com/wesselb/varz}{https://github.com/wesselb/varz}).
Pareto optimisation for the multi-objective optimisation example was performed using \texttt{paretoset} (\href{https://github.com/tommyod/paretoset}{https://github.com/tommyod/paretoset}).
All data used in this study is freely available.
Antarctic station data, containing the station locations used in this study, is available from \href{ftp.bas.ac.uk/src/}{ftp.bas.ac.uk/src/}.
The ERA5 data was downloaded from \href{https://cds.climate.copernicus.eu/cdsapp\#!/dataset/reanalysis-era5-single-levels?tab=overview}{https://cds.climate.copernicus.eu/cdsapp\#!/dataset/reanalysis-era5-single-levels?tab=overview}.
The Antarctic land mask and elevation field was obtained from version 2 of the BedMachine dataset from \href{https://nsidc.org/data/nsidc-0756/versions/2}{https://nsidc.org/data/nsidc-0756/versions/2}.

\paragraph{Ethical Standards}
The research meets all ethical guidelines, including adherence to the legal requirements of the study country.

\paragraph{Author Contributions} Contributions are listed in the order of the author list.
Conceptualisation: T.R.A, W.P.B, S.M, D.J, J.S.H, R.E.T.
Methodology: T.R.A, W.P.B, S.M, J.R, A.V, D.J, R.E.T.
Software: T.R.A, W.P.B.
Data curation: T.R.A.
Visualisation: T.R.A.
Supervision: M.A.L, D.J, J.S.H, R.E.T.
Project administration: D.J, J.S.H.
Funding acquisition: J.S.H.
Writing original draft: T.R.A, W.P.B, S.M, J.R.
Writing – review \& editing: T.R.A, W.P.B, S.M, J.R, A.C-C, A.V, A-L.E, M.A.L, D.J, J.S.H, R.E.T.

\paragraph{Supplementary Material}
An Appendix is included with this submission.

\newpage
\bibliographystyle{apalike}
\bibliography{refs}

\end{Backmatter}



\newpage
\setcounter{page}{1}
\begin{appendix}\appheader

\newcommand{\appendixhead}%
{\noindent \textbf{\Large Appendix for `Environmental Sensor Placement with Convolutional Gaussian Neural Processes'}}

\begin{center}
    \appendixhead
    \vspace{0.1in}
    
    Tom R. Andersson, Wessel P. Bruinsma, Stratis Markou, James Requeima, Alejandro Coca-Castro, Anna Vaughan, Anna-Louise Ellis, Matthew A. Lazzara, Daniel C. Jones, J. Scott Hosking and Richard E. Turner
\end{center}

\vspace{0.25in}

%

\counterwithin{figure}{section}
\counterwithin{table}{section}
\counterwithin{equation}{section}
\renewcommand\thefigure{\thesection\arabic{figure}}
\renewcommand\thetable{\thesection\arabic{table}}

\FloatBarrier
\section{Data considerations}\label{apx.data_considerations}

In this section we provide details on the data sources, preprocessing, and normalisation.

\subsection{Data sources}\label{apx.data.sources}

The daily-averaged temperature reanalysis data was obtained from ERA5 \citep{hersbach_era5_2020}.
The land mask and elevation field was obtained from the BedMachine dataset \citep{morlighem__mathieu_measures_2020}.
Antarctic temperature locations from staffed and automatic weather stations were downloaded from \texttt{ftp.bas.ac.uk/src/}. 

\subsection{Data preprocessing}\label{apx.data.preproc}

The temperature anomaly data and land/elevation auxiliary data were regridded from lat/lon to a Southern Hemisphere Equal Area Scalable Earth 2 (EASE2) grid at \SI{25}{km} resolution and cropping to a size of 280 $\times$ 280.
This centres the data on the South Pole.

Temperature anomalies were computed from the absolute temperature values by first computing the daily-average climatology across 1950-2013 (i.e.~averaging the absolute temperature over time for each day of year, returning a 366 $\times$ 280 $\times$ 280 tensor).
Then, for each day of year, the climatological average was subtracted from the absolute temperature values, returning anomaly values.

\subsection{Data normalisation}\label{apx.data.normalisation}

To aid the training process, we normalised the data before passing it to the ConvGNP and GP models.
The temperature data was normalised from Celsius to a mean of 0 and standard deviation of 1.
The elevation field was normalised from metres to values in $[0, 1]$.
The land mask already took appropriate normalised values in $\{ 0, 1 \}$.
The input coordinates were normalised from metres to take values in $[-1, 1]$.

\FloatBarrier
\section{The ConvGNP model}\label{apx.convgnp}

Here we provide details on the ConvGNP training procedure and architecture. A high-level schematic of the ConvGNP forward-pass is shown in \Cref{fig:foward_pass}. We refer the reader to \citealt{markou_practical_2022} for further model details.

\subsection{Generation of $\D_\tau$ for the training, validation, and test datasets}\label{apx.section.dtau_generation}

Each daily-average training dataset $\D_\tau$ was generated by first drawing the integer number of simulated temperature anomaly context points $N_c \sim \text{Unif} \{5, 6, \dots 500\}$ and target points $N_t \sim \text{Unif} \{3000, 3001, \dots 4000\}$.
Allowing for randomness in $N_c$ encourages the model to learn to deal with both data-sparse and data-rich scenarios.
Using a fairly large number of target points ensures there is sufficient signal for learning the covariance structure of the data while not incurring the computational cost of a very large $N_t$.
Next, given the randomly sampled $N_c$ and $N_t$, the 280 $\times$ 280 ERA5 grid cells were sampled uniformly at random to generate the ERA5 context and target locations, $\bm{X}_\tau\us{(c)}$ and $\bm{X}_\tau\us{(t)}$.

For the training dates, the random seed used for generating $\D_\tau$ is changed every epoch, allowing for an infinitely growing training dataset.
In contrast, for the validation and test dates, fixed random seeds are used so that the $\D_\tau$ are deterministically random.
This ensures the validation and test metrics are deterministic during and after training.

For the test results given in \cref{tab:apx.antarctica-results}, we loop over $N_c \in \{0, 25, 50, \dots 500\}$ and fix $N_t$ to a value of 2,000.
For each setting of $N_c$, we generate test tasks $\D_\tau$ by looping twice over each day in 2018-2019, resulting in 1,458 test tasks per $N_c$.

\subsection{Antarctic surface temperature anomaly ConvGNP training procedure}\label{apx.section.training}

The model was trained on data from 1950-2013.
An Adam optimiser was used with a learning rate of $8 \times 10^{-5}$ and an NLL loss function.
Gradients with respect to the loss were averaged over batches of two datasets.
Validation tasks from 2014-2017 were used for checkpointing the model weights based on the mean NLL over the validation tasks.
In total, the ConvGNP was trained for 170 epochs.
Training took 11.5 days on a Tesla V100 GPU with a simple implementation of the training pipeline.
This could be improved with better computational practices, such as using TensorFlow's graph mode rather than eager mode, which was not supported by the ConvGNP implementation at the time of the experiments.

\subsection{ConvGNP architecture}\label{apx.section.arch}
For the ConvGNP model we use the same architecture as described in \citealt{markou_practical_2022}, except for a few modifications.
The U-Net component of the encoder uses 5x5 convolutional kernels with the following sequence of channel numbers (d.s. = 2x2 downsample layer, u.s. = 2x2 upsample layer):
\begin{equation*}
128 \xrightarrow{\text{d.s.}} 128 \xrightarrow{\text{d.s.}} 128 \xrightarrow{\text{d.s.}} 128 \xrightarrow{\text{u.s.}} 128 \xrightarrow{\text{u.s.}} 128 \xrightarrow{\text{u.s.}} 128.
\end{equation*}
We use 128 basis functions for the covariance-parameterising neural network, $\bm{g}$.
Using 128 channels for each layer in the U-Net means there are no dimensionality bottlenecks that could reduce the actual rank of the output lowrank covariance matrix.
We use bilinear resize operators for the upsampling layers to fix checkerboard artifacts that we encountered when using standard zero-padding upsampling \citep{odena_deconvolution_2016}.
For the internal discretisation density of the model, we used 150 points-per-unit (i.e., a 1$\times$1 square of input space contains 150$\times$150 internal discretisation points).
We chose 150 points-per-unit to be close to the density of the ERA5 data in normalised coordinates, which is 140$\times$140 in a 1$\times$1 square of input space.\footnote{Note, finer resolution context data would motivate the use of higher internal discretisation densities.}

The hyperparameter settings above construct a ConvGNP with 4.16 million learnable parameters.
In addition, the choices for the U-Net filter size and internal discretisation density results in a receptive field of over \SI{1500}{km}.
In other words, context observations can influence target predictions in the Gaussian predictive distribution up to roughly \SI{750}{km} in either direction of the $\bm{x}_1$- or $\bm{x}_2$-dimensions.

\subsection{ConvGNP input data}\label{apx.section.convgnp_input}

The ConvGNP receives two context sets as input.
The first contains observations of the simulated ERA5 daily-average temperature anomaly.
The second contains a set of 6 gridded auxiliary and metadata variables. These are: elevation, land mask, $\cos(\text{day of year})$, $\sin(\text{day of year})$, $x_1$, and $x_2$.
The $\cos(\text{day of year})$ and $\sin(\text{day of year})$ inputs, where the day of year is normalised between 0 and $2\pi$, together define a circular variable that rotates once per year.
This informs the model at what time of year $\D_\tau$ corresponds to, helping with learning seasonal variations in the data.
The $x_1$ and $x_2$ gridded fields inform the model where in input space the data corresponds to.
The gridded auxiliary fields that vary over input space are crucial for allowing the ConvGNP to model spatial non-stationarity.
This is because they break the U-Net's translation equivariance property.
Future work could explore using \textit{learnable} input auxiliary channels, as in \citealt{addison_machine_2022}, which could lead to richer non-stationarity at the cost of a greater spatial overfitting risk.

\subsection{The SetConv encoder enables fusing data sources with multiple modalities and missing values}\label{apx.subsection.setconv}

The SetConv encoder \citep{gordon_convolutional_2020}, which fuses the context sets into a single gridded tensor, enables the ConvGNP to model 1) missing data, 2) multiple data streams, and 3) both gridded and off-grid data modalities.
This is achieved, in part, through the use of a `density channel' for each context set \citep{gordon_convolutional_2020}.
The density channel measures the density of context points by placing a small Gaussian basis function of unit amplitude at each observation location, such that the density channel takes values close to zero away from observations.
Output $y$-values of context points are encoded in a similar fashion, but the amplitude of the Gaussian basis function is weighted by the value of $y$.
These intermediate functional representations are then discretised onto the model's internal grid to yield the density channels and ($N$-dimensional) `data channels' in the gridded encoding.
This set-up allows the model to distinguish between the case where an observation is made with a $y$-value of $0$ (data channel is zero but density channel is non-zero), and the case where there is no observation available (both the data and density channels are zero).

\begin{figure}[t]
    \centering
    \includegraphics[scale=0.9]{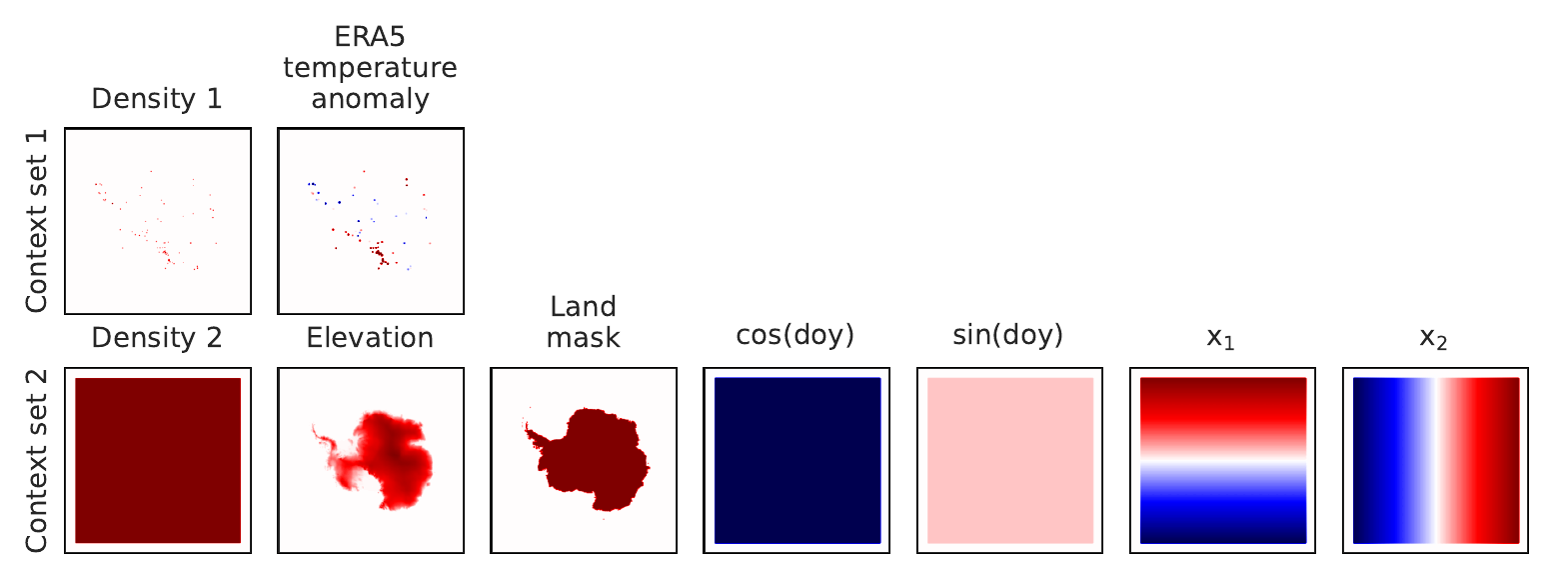}
    \caption{Example output of the ConvGNP's SetConv encoder. The SetConv was passed a context set with ERA5 temperature anomaly at Antarctic station locations.}
    \label{fig:apx.setconv}
\end{figure}

\Cref{fig:apx.setconv} shows the channels of a gridded encoding, which are splayed out to highlight the two input context sets provided to the ConvGNP in this study. 
The density channel for the first context set pinpoints the scattered, point-based temperature anomaly observation locations.
The second density channel pertains to the gridded auxiliary context set and thus takes a constant value within the region of data.
While in this setting the second gridded context set contained auxiliary variables with no missing data, the SetConv can represent missing data with gridded variables as well as non-gridded variables.
For example, missing satellite observations due to cloud cover would be captured by patterns of zeros in the density channel.
The density channel can thus be seen as a kind of \textit{missing data channel}, where missing data (e.g.~due to sensor malfunction, clouds, or the absence of sensing equipment) is captured with density values close to zero.
Therefore, the SetConv encoder equips geospatial deep learning models with an ability to handle missing data, which is an important problem in many application areas \citep{mitra_learning_2023}.
However, the degree to which the model can learn to respond to missing data appropriately depends on a training scheme with sufficient examples of missing data, as discussed in \Cref{section.discussion}.


\newpage
\FloatBarrier
\section{Gaussian Process benchmarks}\label{apx.section.benchmarks}

Here we provide details on the GP baseline kernels and hyperparameter fitting procedure.
All GPs were implemented using the Python package \texttt{stheno} (https://github.com/wesselb/stheno) and optimised using the Python package \texttt{varz} (https://github.com/wesselb/varz).







\subsection{Gibbs GP}\label{apx.benchmarks.gibbs}

The Gibbs kernel \citep{gibbs} is a non-stationary generalisation of the EQ kernel.
In the $\bm{x} \in \mathbb{R}^2$ case, the covariance function is given by:
\begin{equation}\label{eq:gibbs}
    k_\text{Gibbs}(\bm{x}, \bm{x}') = \sigma^2 \prod_{i=1}^2 \bigg( \frac{2 l_i(\bm{x}) l_i(\bm{x}')}{l_i(\bm{x})^2 + l_i(\bm{x}')^2} \bigg)^{1/2} \exp \bigg( -\sum_{i=1}^2 \frac{(x_i - x_i')^2}{l_i(\bm{x})^2 + l_i(\bm{x}')^2} \bigg),
\end{equation}
where $\sigma^2$ is the variance, and length scale functions $l_1(\bm{x})$ and $l_2(\bm{x})$ dictate the length scales in the $x_1$- and $x_2$-directions.
We parameterise the length scale functions $l_i(\bm{x})$ as a weighted sum of $M$ regularly placed Gaussian basis functions,
\begin{equation}
l_i (\bm{x}) = \sum_{m=1}^M \theta_{i,m} \exp\bigg(-\frac{(\bm{x}-\bm{x}^{(\mu)}_m)^T (\bm{x}-\bm{x}^{(\mu)}_m)}{2\lambda^2}\bigg),
\end{equation}
where the $\theta_{i,m}$ are the constrained-positive weights of basis function $m$ for input dimension $i$, and the basis functions are placed with the $\bm{x}^{(\mu)}_m$ on a 100$\times$100 grid spanning the input space.
The basis function length scale $\lambda$ is kept fixed and equal to the spacing between basis functions.
We note that the basis function spacing controls how quickly the length scale functions can vary, and is a fixed (untrainable) hyperparameter. 
Too many basis functions can lead to overfitting, while too few can lead to underfitting.
We tried different settings and chose a 100$\times$100 grid as the most performant.

We train the parameters $\{\bm{\theta}_1, \bm{\theta}_2, \sigma\}$, along with the other hyperparameters, using gradient descent on the negative log marginal likelihood (NLML) using an Adam optimiser with learning rate of $5 \times 10^{-3}$ and a batch size of 10.
We used 1950-2013 as a training period, subsampling the dates by a factor of 3, and sampling 500 random context locations for each of the training tasks \Cref{apx.section.dtau_generation}.
Training was halted after the NLL on validation data spanning 2014--2017 did not improve for 5 epochs.

\begin{figure}[H]
    \centering
    \includegraphics[trim=1.5cm .5cm 0cm .5cm,clip,scale=.9]{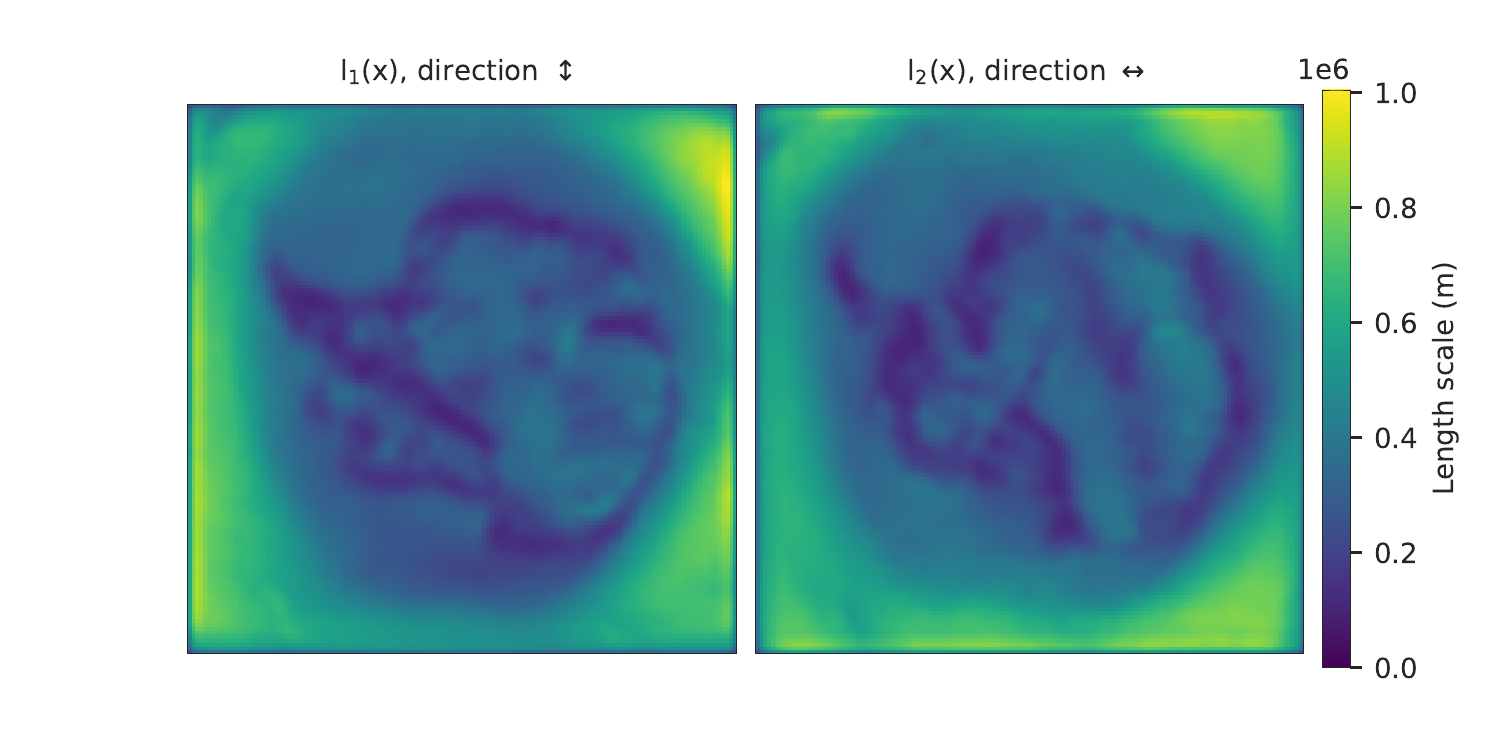}
    \caption{Learned Gibbs GP length scale functions, $l_1(\bm{x})$ and $l_2(\bm{x})$.}
    \label{fig:apx.gibbs_l_x}
\end{figure}

\Cref{fig:apx.gibbs_l_x} shows the trained length scale functions $l_1(\bm{x})$ and $l_2(\bm{x})$, revealing interesting detail such as very low correlation length scales perpendicular to the coastline.

\subsection{Exponentiated quadratic and rational quadratic GPs}\label{apx.benchmarks.eq}

We also include more simplistic GP baselines using non-isotropic exponentiated quadratic (EQ) and rational quadratic (RQ) kernels, which are both stationary prior covariance functions (unlike the Gibbs kernel).
The non-isotropic EQ kernel is:
\begin{equation}\label{eq:eq}
k_\text{EQ}(\mathbf{x},\mathbf{x}^\prime) = \sigma^2 \exp\left(-\frac{(x_1 - x_1^\prime)^2}{2\ell_1^2} -\frac{(x_2 - x_2^\prime)^2}{2\ell_2^2}\right),
\end{equation}
where $\sigma^2$ is the variance, and $\ell_1$ and $\ell_2$ are the length scales in each input dimension.
The non-isotropic RQ kernel is:
\begin{equation}\label{eq:rq}
k_{\text{RQ}}(\mathbf{x}, \mathbf{x}^\prime) = \sigma^2\left(1+\frac{(x_1 - x_1^\prime)^2}{2\alpha\ell_1^2} + \frac{(x_2 - x_2^\prime)^2}{2\alpha\ell_2^2}\right)^{-\alpha},
\end{equation}
where $\alpha$ is the shape parameter, controlling the smoothness of the kernel.
The RQ kernel can be seen as an infinite sum of EQ kernels with different length scales \citep{rasmussen_gaussian_2004}.

We fit the EQ and RQ GP hyperparameters using the L-BFGS-B algorithm on a batch of 730 dates sampled randomly from 1950-2013.
The EQ and RQ GPs are thus exposed to fewer training tasks.
However, these models only have a few parameters each.
We found that increasing the training set size did not yield improved performance.

\newpage
\FloatBarrier
\section{Non-stationarity in the ConvGNP}\label{apx.section.nonstationarity}
\begin{figure}[h]
    \centering
    \includegraphics[scale=.9]{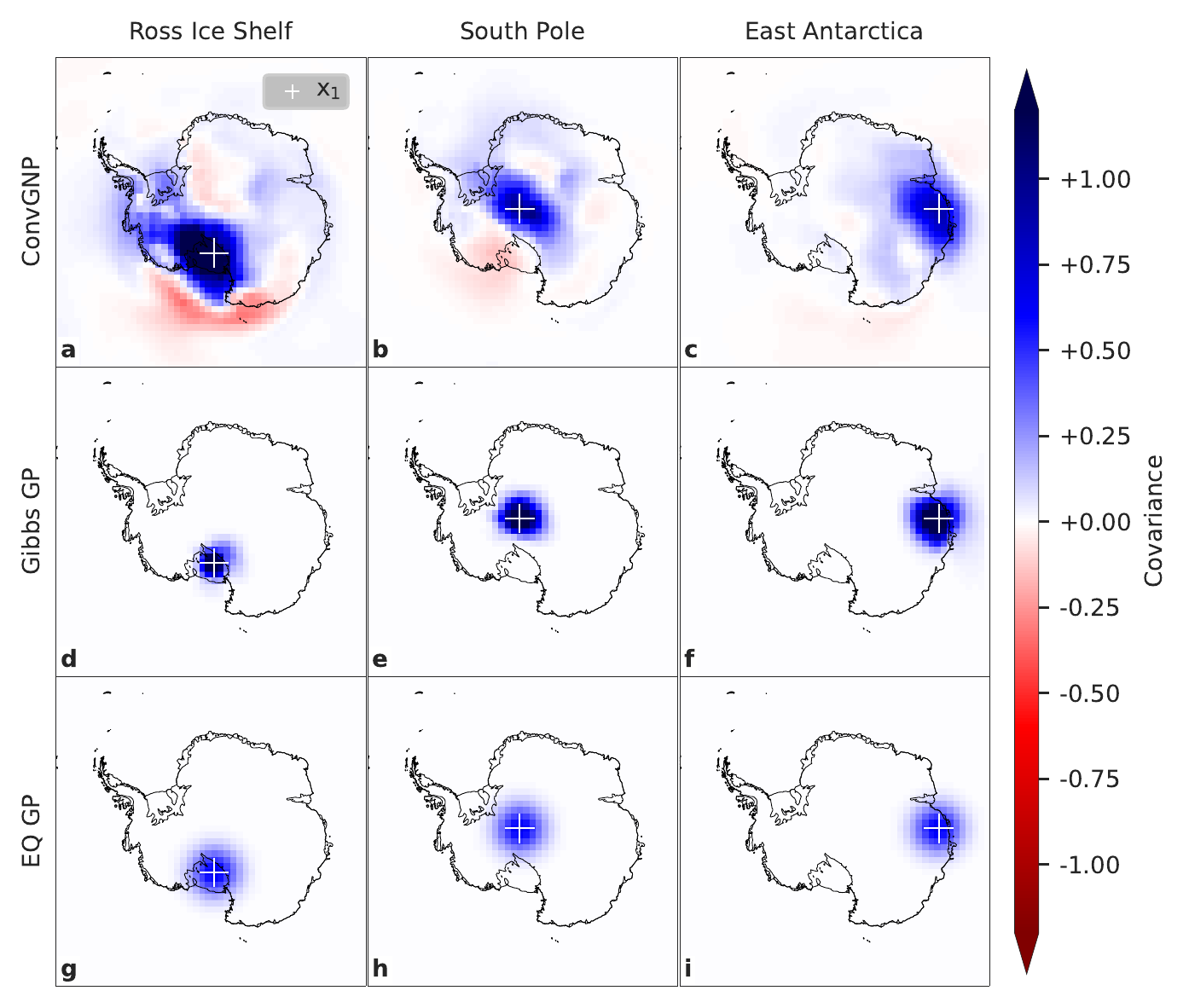}
    \caption{\textbf{The ConvGNP learns spatially-varying covariance structure.}
    Heatmaps showing the prior covariance function, $k(\bm{x}_1, \bm{x}_2)$, with $\bm{x}_1$ fixed at the white plus location and $\bm{x}_2$ varying over the grid. Plots are shown for three different $\bm{x}_1$-locations (the Ross Ice Shelf, the South Pole, and East Antarctica) and the three models (ConvGNP, Gibbs GP, and EQ GP). The ConvGNP's day of year input was the 1st of June.}
    \label{fig:apx.cov}
\end{figure}

\begin{figure}[h]
    \centering
    \includegraphics[scale=1]{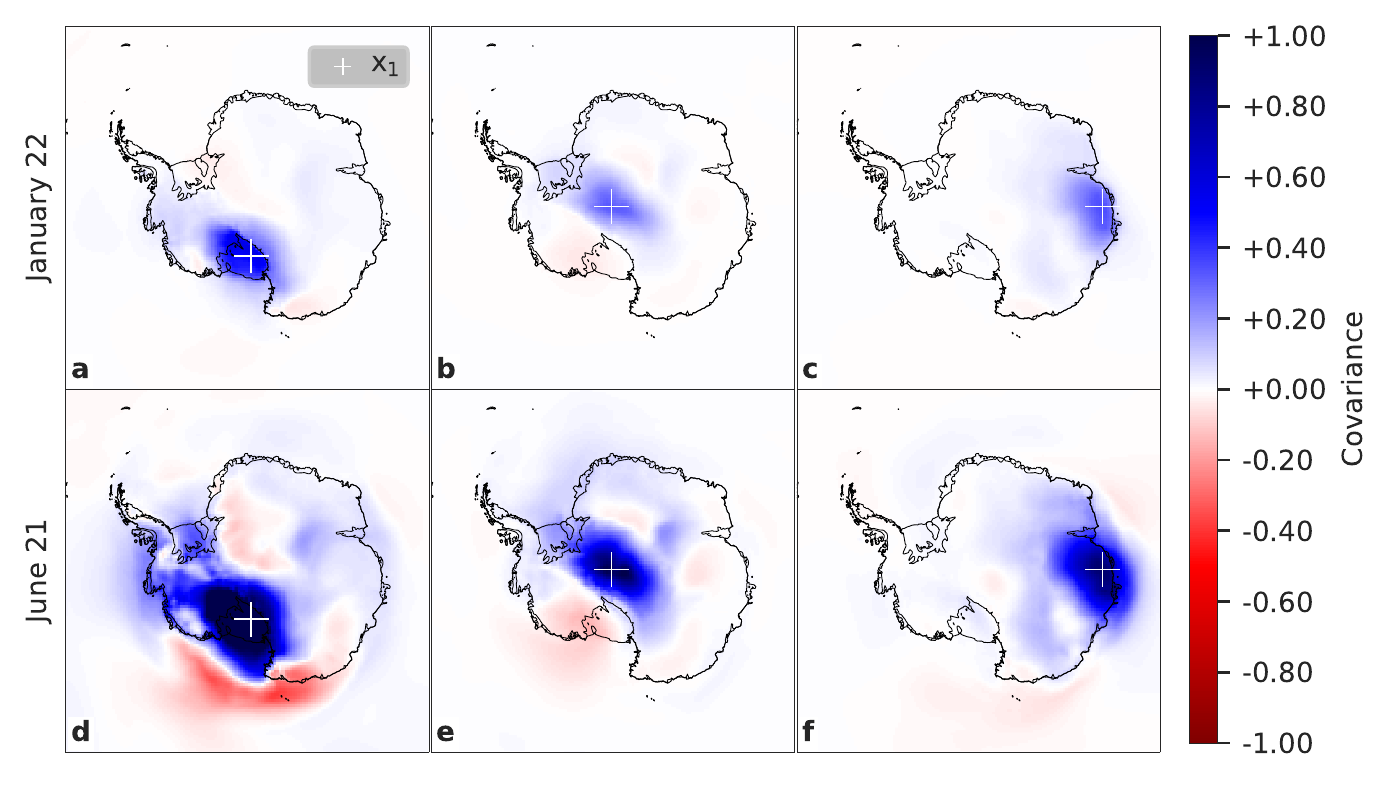}
    \caption{\textbf{The ConvGNP learns seasonally-varying covariance.}
    Heatmaps showing the ConvGNP's prior covariance function $k(\bm{x}_1, \bm{x}_2)$, as in \Cref{fig:cov}a-c, but for two times of the year: midsummer (Jan 22nd) and midwinter (June 21st). This shows that the ConvGNP has learned a prior covariance function with non-stationarity over day of year.
    }
    \label{fig:apx.cov_time}
\end{figure}

\begin{figure}[h]
    \centering
    \includegraphics[scale=1]{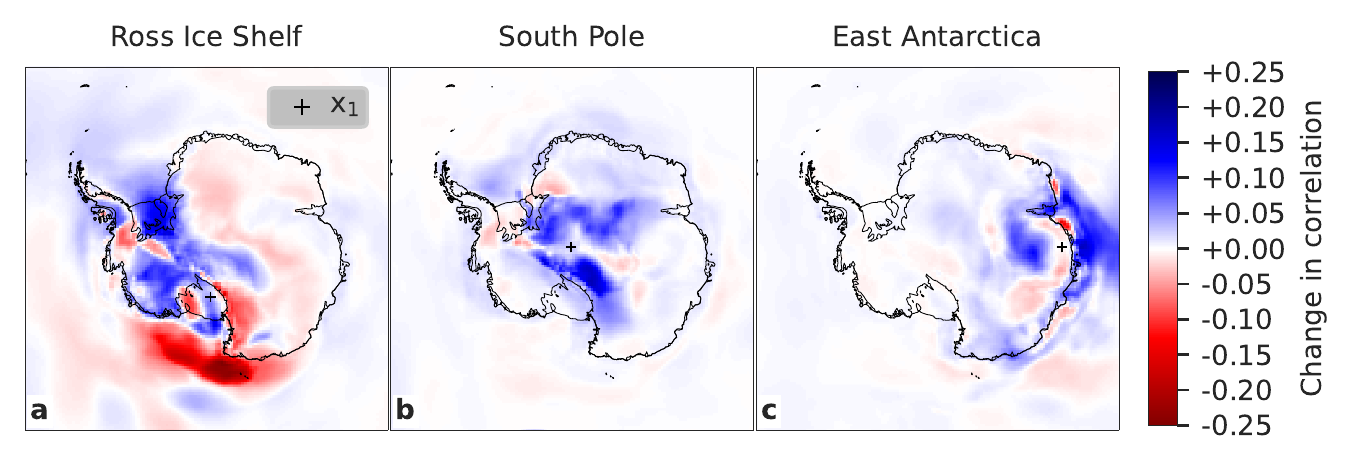}
    \caption{\textbf{The ConvGNP learns seasonal changes in spatial correlations.} Change in prior correlation from the ConvGNP from the Austral Midsummer (Jan 22nd) to Midwinter (Jun 21st). The correlation $\rho$ was computed from the covariance using $\rho = k(\bm{x}_1, \bm{x}_2) / \sqrt{k(\bm{x}_1, \bm{x}_1) k(\bm{x}_2, \bm{x}_2)}$, with $\bm{x}_1$ fixed at the black plus location and $\bm{x}_2$ varying over the grid. Plot shows summer minus winter.}
    \label{fig:apx.corr_diff}
\end{figure}

The ConvGNP learns richer spatial covariance structure than the GP baselines (\Cref{fig:apx.cov}).
Further, while our implementation of the ConvGNP only models correlations over 2D space (i.e.~modelling time independently), the model can leverage the day of year auxiliary inputs to learn seasonal non-stationarity in the data (\Cref{fig:apx.cov_time}).
We note that the main changes in the ConvGNP's covariance from summer to winter are caused by changes in the magnitudes of the marginal variances (temperature anomalies take more extreme values in winter).
However, the spatial correlations also change (\Cref{fig:apx.corr_diff}).
For example, the Ross Ice Shelf site becomes less correlated with the Southern Ocean (\Cref{fig:apx.corr_diff}a), the South Pole becomes more correlated with the surrounding region (\Cref{fig:apx.corr_diff}b), and the East Antarctica site becomes more correlated with the Southern Ocean (\Cref{fig:apx.corr_diff}c).


\newpage
\FloatBarrier
\section{Additional test set results}

In this section we provide further details on the models' test set performance.

\subsection{Overall performance metrics}\label{apx.test_results}

\Cref{tab:apx.antarctica-results} shows the test set results averaged over $N_c \in \{0, 25, 50, \dots 500\}$.
The ConvGNP outperforms the three GP baselines with statistical significance across all three metrics (the normalised joint NLL, the marginal NLL, and the RMSE).

\begin{table}[H]
    \centering
    \small
    \caption{Performance on test tasks from the period $2018$--$2019$, using $N_c \in \{0, 25, 50, \dots 500\}$. Errors indicate standard errors. For each metric, lower is better. Significantly best results in bold.}
    \label{tab:apx.antarctica-results}
    \scshape
    \begin{tabular}{rcccc}
        \toprule
        Metric & ConvGNP & Gibbs GP & RQ GP & EQ GP \\ \midrule
        Normalised joint NLL &
            $\bm{-0.61} {\scriptstyle \pm 0.00}$ & $-0.42 {\scriptstyle \pm 0.00}$ & $-0.23 {\scriptstyle \pm 0.00}$ & $\hphantom{-}0.00 {\scriptstyle \pm 0.00}$ \\
        Marginal NLL &
            $\bm{-0.19} {\scriptstyle \pm 0.01}$ & $\hphantom{-}0.30 {\scriptstyle \pm 0.01}$ & $\hphantom{-}0.47 {\scriptstyle \pm 0.01}$ & $\hphantom{-}0.54 {\scriptstyle \pm 0.01}$ \\
        RMSE ($^{\circ}$C) & 
            $\hphantom{\bm{-}}\bm{1.54} {\scriptstyle \pm 0.01}$ & $\hphantom{-}1.84 {\scriptstyle \pm 0.01}$ & $\hphantom{-}1.63 {\scriptstyle \pm 0.01}$ & $\hphantom{-}1.72 {\scriptstyle \pm 0.01}$\\
        \bottomrule
    \end{tabular}
    \vspace{-3mm}
\end{table}

\subsection{Marginal calibration and sharpness}\label{apx.marginal_calib}

The calibration and sharpness of probabilistic prediction systems are key performance indicators \citep{gneiting_probabilistic_2007}.
To assess these two quantities, we generated test tasks by subsampling the test dates (2018--2019) by a factor of 30 to obtain 25 dates.
We then followed the same procedure to generate test tasks as in \Cref{apx.section.dtau_generation}--looping over $N_c \in \{0, 25, 50, \dots 500\}$ with $N_t=2000$, resulting in 50,000 marginal predictions per $N_c$ for each model.

\begin{figure}[h]
    \centering
    \includegraphics[scale=1]{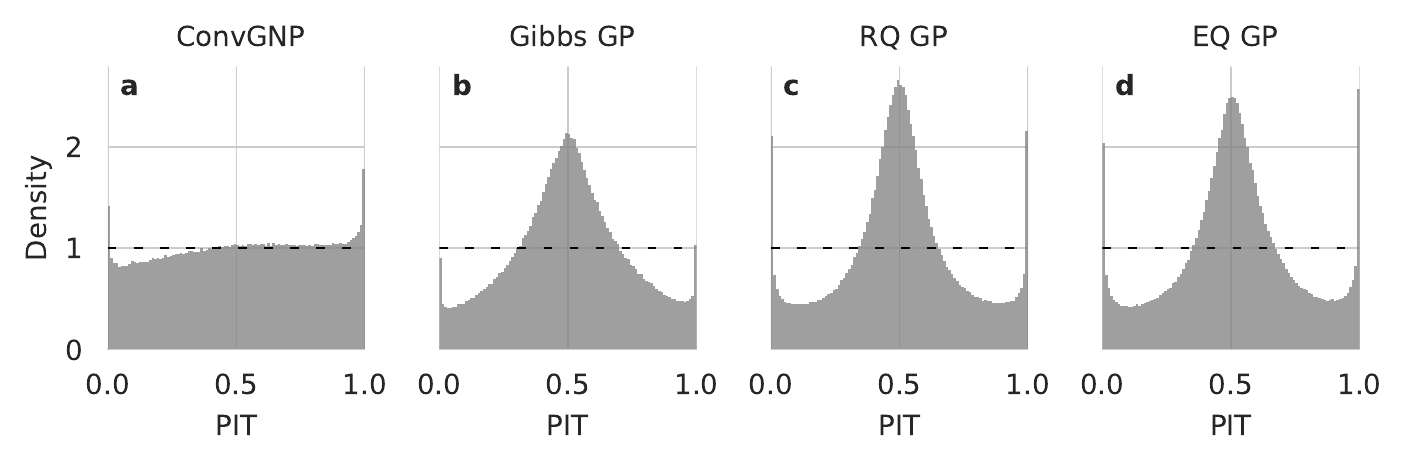}
    \caption{\textbf{The ConvGNP produces the most well-calibrated marginal predictions.} Probability integral transform (PIT) histograms evaluated on 25 dates from the test years (2018--2019). The PIT is defined as the cumulative distribution function (CDF) of the model's marginal distribution evaluated at the true $y$-values. A black dashed line shows the ideal uniform distribution (corresponding to perfect calibration).}
    \label{fig:apx.marginal_calib}
\end{figure}

The calibration of a model's marginal distributions can be assessed using the probability integral transform (PIT).
The PIT is defined as the cumulative distribution function (CDF) of the model's marginal distribution evaluated at the true observed value of a particular target point.
If a model has perfect calibration, the histogram of PIT values is uniform \citep{gneiting_probabilistic_2007}. 
When aggregating across all test tasks (with varying $N_c$ values), the ConvGNP's marginal distributions are much better calibrated than the GP baselines, coming closer to the ideal uniform distribution of PIT values (\Cref{fig:apx.marginal_calib}).

\begin{figure}[h]
    \centering
    \includegraphics[scale=1]{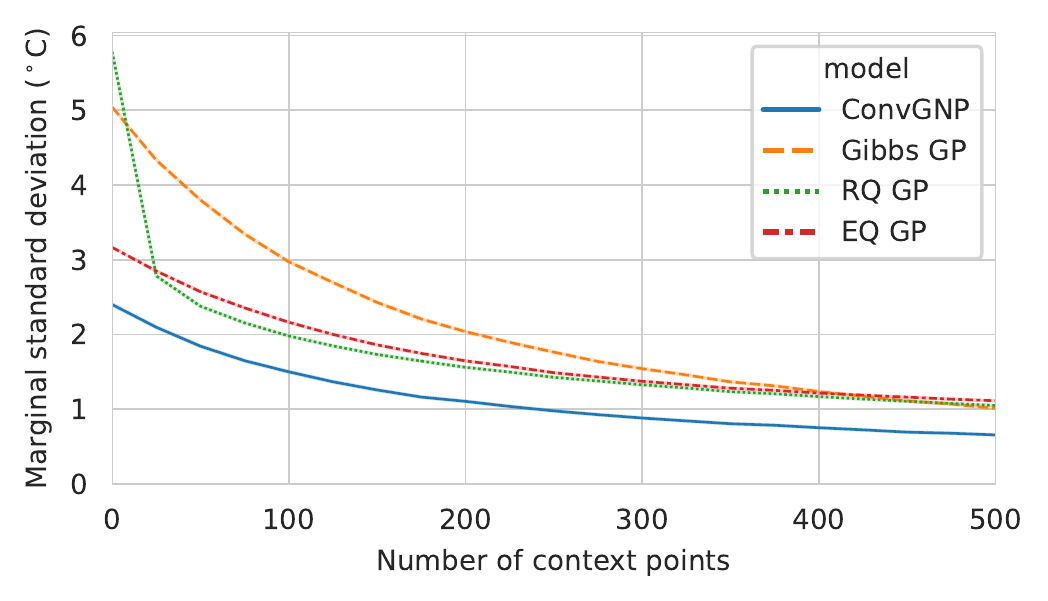}
    \caption{\textbf{The ConvGNP produces the sharpest marginal predictions.} Mean standard deviation of the models' marginal Gaussian distributions versus number of context points, $N_c$. 50,000 standard deviation values were used per $N_c$, derived from 25 dates in the test years (2018--2019). Error bars are standard errors.}
    \label{fig:apx.marginal_sharpness}
\end{figure}

The sharpness (i.e.~degree of certainty) of probabilistic predictions must also be considered alongside calibration; a key goal for probabilistic prediction systems is to maximise sharpness subject to good calibration \citep{gneiting_probabilistic_2007}.
We assess marginal distribution sharpness by plotting the standard deviation of the univariate Gaussian marginals against $N_c$.
The ConvGNP makes substantially more confident predictions than the GP baselines across all values of $N_c$ \Cref{fig:apx.marginal_sharpness}.
The GP baselines tend to make uninformative predictions with large uncertainty, explaining why their PIT values cluster around 0.5 (\Cref{fig:apx.marginal_calib}b--c).

\FloatBarrier
\section{Sensor placement acquisition functions}\label{apx.section.acquisition_functions}

Here we expand upon and mathematically define each acquisition function used for sensor placement. \\

\subsection{Model-based uncertainty reduction acquisition functions}\label{apx.section.model_based_acquisition_functions}


\vspace{1mm}
\noindent \textbf{JointMI}: Expected reduction in joint entropy over targets after appending the query sensor to the context set $C$.
\begin{align}
    \alpha(\bm{x}\us{(s)}_i, \tau) &= H(\bm{y}_\tau\us{(t)} | C_\tau) - 
\E_{\pi(y_{\tau, i}\us{(s)})} \bigg[ H(\bm{y}_\tau\us{(t)}|C_\tau, \bm{x}_{i}\us{(s)}, y_{\tau, i}\us{(s)}) \bigg] \label{apx.eq:alpha_mi} \\ 
     &= H(\bm{y}_\tau\us{(t)} | C_\tau) - \int 
\pi(y_{\tau, i}\us{(s)} ; C_\tau) H(\bm{y}_\tau\us{(t)}|C_\tau, \bm{x}_{i}\us{(s)}, y_{\tau, i}\us{(s)}) \diff y_{\tau, i}\us{(s)} \\ 
    &\stackrel{(a)}{\approx} H(\bm{y}_\tau\us{(t)} | C_\tau) - H(\bm{y}_\tau\us{(t)}|C_\tau, \bm{x}_{i}\us{(s)}, \bar{y}_{\tau, i}\us{(s)}) \label{apx.eq:entropy_approx}\\
    &\stackrel{(b)}{\approx} c_\tau - H(\bm{y}_\tau\us{(t)}|C_\tau, \bm{x}_{i}\us{(s)}, \bar{y}_{\tau, i}\us{(s)}), \\
    &\approx c_\tau - \frac{1}{2} \log ( (2\pi e)^2 |\bm{K}|), \\
    &\approx c_\tau - \frac{1}{2} \log |\bm{K}|,\label{eq:apx.jointmi}
\end{align}
where $c_\tau$ is a constant and $\bm{K}$ is the model's covariance matrix at the target locations after conditioning on the imputed query sensor observation, $(\bm{x}_{i}\us{(s)}, \bar{y}_{\tau, i}\us{(s)})$.
In (a) we approximate the intractable expectation integral over the entropy term $H(\bm{y}_\tau\us{(t)}|C_\tau, \bm{x}_{i}\us{(s)}, y_{\tau, i}\us{(s)})$ with a simple substitution of the model's mean prediction $\bar{y}_{\tau, i}\us{(s)}$ at query location $\bm{x}\us{(s)}_i$.
In (b) we use the fact that $H(\bm{y}_\tau\us{(t)} | C_\tau)$ depends only on 
$\tau$ and not $\bm{x}\us{(s)}_i$.
Thus, this placement criterion is equivalent to minimising the entropy over $\bm{y}_\tau\us{(t)}|C_\tau, \bm{x}_{i}\us{(s)}, \bar{y}_{\tau, i}\us{(s)}$ by minimising the determinant of the covariance matrix $\bm{K}$.

This placement criterion may be hindered by approximating the expectation over the query observation $y_{\tau, i}\us{(s)}$ by imputing with the model's mean prediction $\bar{y}_{\tau, i}\us{(s)}$ in Equation~\ref{apx.eq:entropy_approx}. 
Instead, a better scheme would draw samples from the model's marginal distribution over $y_{\tau, i}\us{(s)}$ to estimate the expectation using Monte Carlo sampling, which may better predict the actual reduction in entropy upon conditioning on the true observation.
However, Monte Carlo sampling linearly increases the cost of evaluating the acquisition function.
Future work should quantify the performance boost from switching to this sampling procedure.




\vspace{1mm}
\noindent \textbf{MarginalMI}: Expected reduction in marginal entropy over targets upon conditioning on the query sensor.
Equivalent to that of Equation~\ref{eq:apx.jointmi}, but setting the off-diagonal covariances in the model's output Gaussian distribution to zero when evaluating the entropy term $H(\bm{y}_\tau\us{(t)}|C_\tau, \bm{x}_{i}\us{(s)}, y_{\tau, i}\us{(s)})$.
Using the resulting independence of the individual marginal distributions after step (b) in Equation A.5 leads to:
\begin{align}
    \alpha(\bm{x}\us{(s)}_i, \tau) &\approx c_\tau - \sum_{j=1}^{N_t} H(y_{j, \tau}\us{(t)}|C_\tau, \bm{x}_{i}\us{(s)}, \bar{y}_{\tau, i}\us{(s)}), \\
    &\approx c_\tau - \sum_{j=1}^{N_t} \frac{1}{2}\log(2 \pi e \sigma_{\tau, j}^2), \\
    &\approx c_\tau - \sum_{j=1}^{N_t} \log(\sigma_{\tau, j}^2),
\end{align}
where $\sigma_{\tau, j}^2$ is the variance of the model's marginal Gaussian distribution of target point $j$ at time $\tau$ after conditioning on the imputed query sensor observation, $(\bm{x}_{i}\us{(s)}, \bar{y}_{\tau, i}\us{(s)})$.
In other words, the acquisition function is the negative sum of the marginal Gaussian entropies at each target location after adding the query sensor to the context set.
After this acquisition function is averaged over time steps in Equation~\ref{eq:alpha_avg}, the placement criterion amounts to minimising the mean log-variance over time and target locations.

\vspace{1mm}
\noindent \textbf{DeltaVar}: Expected reduction in mean marginal variance over targets upon conditioning on the query sensor.
Following the same expectation approximation as \texttt{JointMI} and \texttt{MarginalMI} we arrive at:
\begin{align}
    \alpha(\bm{x}\us{(s)}_i, \tau) &\approx c_\tau - \frac{1}{N_t}\sum_{j=1}^{N_t} \sigma_{\tau, j}^2.
\end{align}
The main difference with the \texttt{MarginalMI} acquisition function is that \texttt{DeltaVar} minimises the absolute marginal variances rather than the log marginal variances.

\subsubsection{Model-based uncertainty reduction acquisition functions from the sensor placement experiment}

\begin{figure}[h]
    \vspace{-3mm}
    \centering
    \includegraphics[trim=0cm 0cm 0cm 0cm,clip,scale=.85]{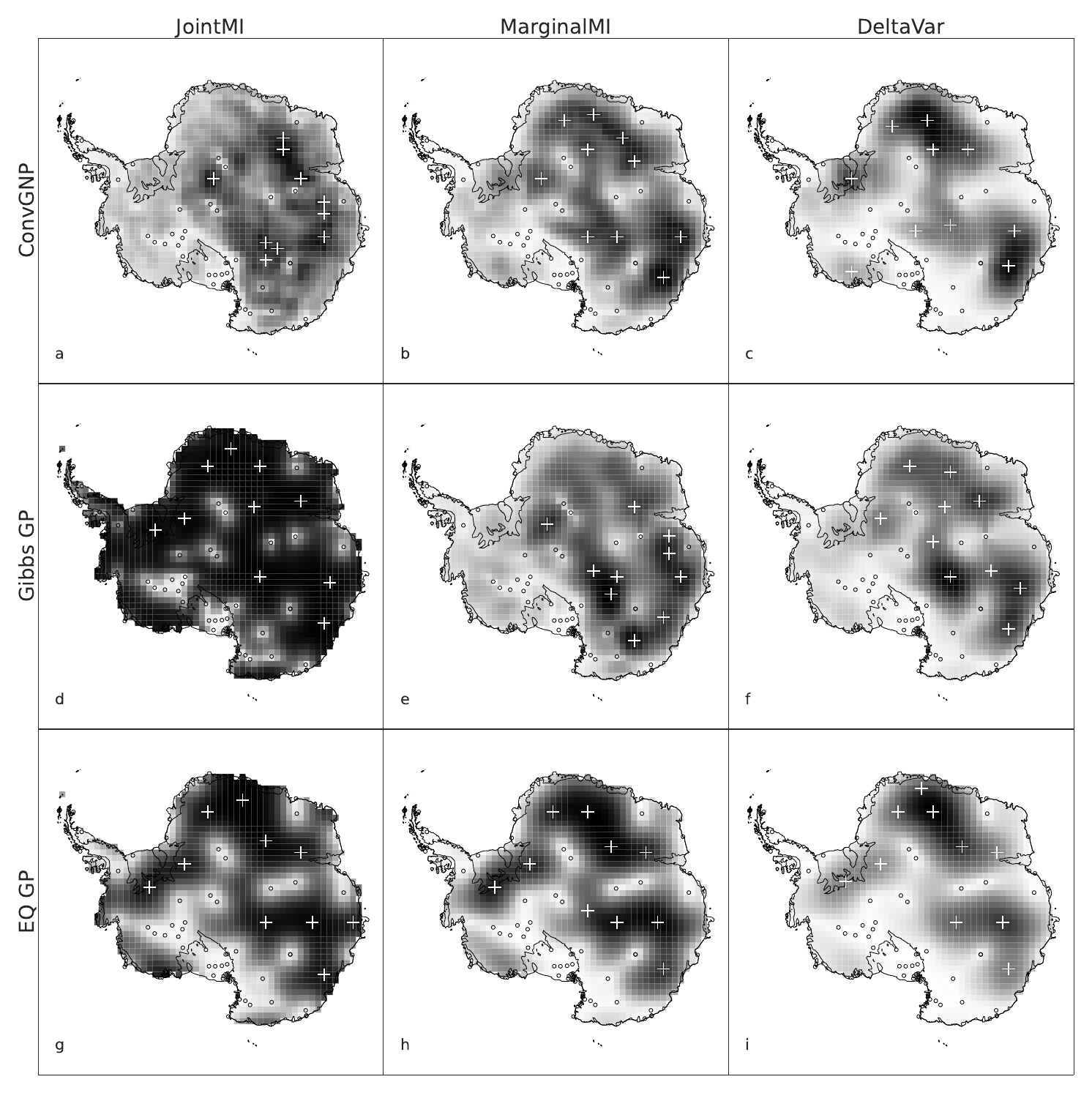}
    \vspace{-3mm}
    \caption{\textbf{Acquisition functions and sensor placements for all three models.} Maps of acquisition function values $\alpha(\bm{x}\us{(s)}_i)$ for the initial $k=1$ greedy iteration. The initial context set $\bm{X}^{(c)}$ is derived from real Antarctic station locations (indicated by black crosses). Running the sensor placement algorithm for $K=10$ sensor placements results in the proposed sensor placements $\bm{X}^*$ (indicated by pluses).}
    \label{fig:apx.sp_all}
    \vspace{-2mm}
\end{figure}

\Cref{fig:apx.sp_all} plots heatmaps of the three model-based acquisition functions (\texttt{JointMI}, \texttt{MarginalMI}, and \texttt{DeltaVar}) at the first greedy iteration for the ConvGNP, Gibbs GP, and EQ GP.
Also shown are the $K=10$ proposed sensor placements with the criterion of maximising these acquisition functions.

\newpage
\subsection{Baseline acquisition functions}

\vspace{1mm}
\noindent \textbf{ContextDist}: Euclidean distance from the closest context point,
\begin{equation}
    \alpha(\bm{x}\us{(s)}_i, \tau) = \min \{||\bm{x}\us{(s)}_i - \bm{x}_{\tau,1}\us{(c)}||_2, ..., ||\bm{x}\us{(s)}_i - \bm{x}_{\tau,N_c}\us{(c)}||_2\},
\end{equation}

\vspace{1mm}
\noindent \textbf{Random}: Uniform at random in [0, 1],
\begin{equation}
    \alpha(\bm{x}\us{(s)}_i, \tau) = u_{\tau, i} \quad \text{where} \quad u_{\tau, i} \sim \text{Unif} (0, 1).
\end{equation}

\noindent Maximising this random acquisition function results in placements that are sampled uniformly from the search points $\bm{X}\us{(s)}$.

\subsection{Oracle acquisition functions}\label{apx.section.oracle_acquisition_functions}
Let a performance metric $\gamma$ take in the probability distribution over targets output by prediction map $\pi$ and the true target values $\bm{y}_\tau\us{(t)}$.
Considering only the 1D context set corresponding to observations of the target variable for notational simplicity, the oracle acquisition functions are:
\begin{align}
     \alpha_\text{oracle}(\bm{x}\us{(s)}_i, \tau) &= \gamma( \pi(\bm{y}_\tau\us{(t)}; \bm{X}_\tau\us{(c)}, \bm{y}_\tau\us{(c)}, \bm{X}_\tau\us{(t)}), \bm{y}_\tau\us{(t)}) - \gamma( \pi(\bm{y}_\tau\us{(t)}; \{\bm{X}_\tau\us{(c)}, \bm{x}\us{(s)}_i\},  \{\bm{y}_\tau\us{(c)}, y\us{(s)}_{\tau, i}\}, \bm{X}_\tau\us{(t)}), \bm{y}_\tau\us{(t)}), \\
     &= c_\tau - \gamma( \pi(\bm{y}_\tau\us{(t)}; \{\bm{X}_\tau\us{(c)}, \bm{x}\us{(s)}_i\},  \{\bm{y}_\tau\us{(c)}, y\us{(s)}_{\tau, i}\}, \bm{X}_\tau\us{(t)}), \bm{y}_\tau\us{(t)}),
\end{align}
\noindent which is the decrease in performance metric (assuming lower is better) induced by concatenating the true observation $(\bm{x}\us{(s)}_i, y\us{(s)}_{\tau, i})$ to the context set.
$\alpha_\text{oracle}(\bm{x}\us{(s)}_i, \tau)$ is then averaged over $\tau$ as in \Cref{eq:alpha_avg} to obtain $\alpha_\text{oracle}(\bm{x}\us{(s)}_i)$.



\subsection{Comment on the dependence on context observations in the acquisition functions}\label{apx.section.ydependence}

The posterior covariance function of a vanilla GP depends only on the input locations $\bm{X}\us{(c)}$ of the context set, not the observed values $\bm{y}\us{(c)}$.
Consequently, the \texttt{JointMI}, \texttt{MarginalMI}, and \texttt{DeltaVar} placement methods will depend only on the input locations.
This behaviour is noted by \citealt{mackay_information-based_1992} for a Bayesian linear regression model with a Gaussian prior, which is a special case of a GP \citep{rasmussen_gaussian_2004}.
This could be seen as an inflexible limitation of GPs; they cannot augment their posterior correlation structure based on the $y$-values observed at the $\bm{x}$-locations. 
For example, if an extreme $y$-value is observed in the context set, a GP posterior cannot become more uncertain, which may be a desirable characteristic.
By construction, the ConvGNP is a non-linear map from context sets to GPs, which means that the whole GP, including the covariance, can depend on every aspect of the context set, including the $y$-values.
The ConvGNP's $y$-dependence necessitates the expectation integral over the unobserved query $y$-value in Equation~\ref{apx.eq:alpha_mi}, as well as the averaging over multiple time steps for the uncertainty-based acquisition functions in Equation~\ref{eq:alpha_avg}.
Neither of these steps are necessary for the GP baselines since their covariance is independent of the $y$-values.

\newpage

\FloatBarrier
\section{Oracle sensor placement results}\label{apx.section.oracle_placement_results}

\begin{figure}[h]
    \centering
    \includegraphics[trim=1.2cm 1cm 1.2cm 0cm,clip,scale=1.1]{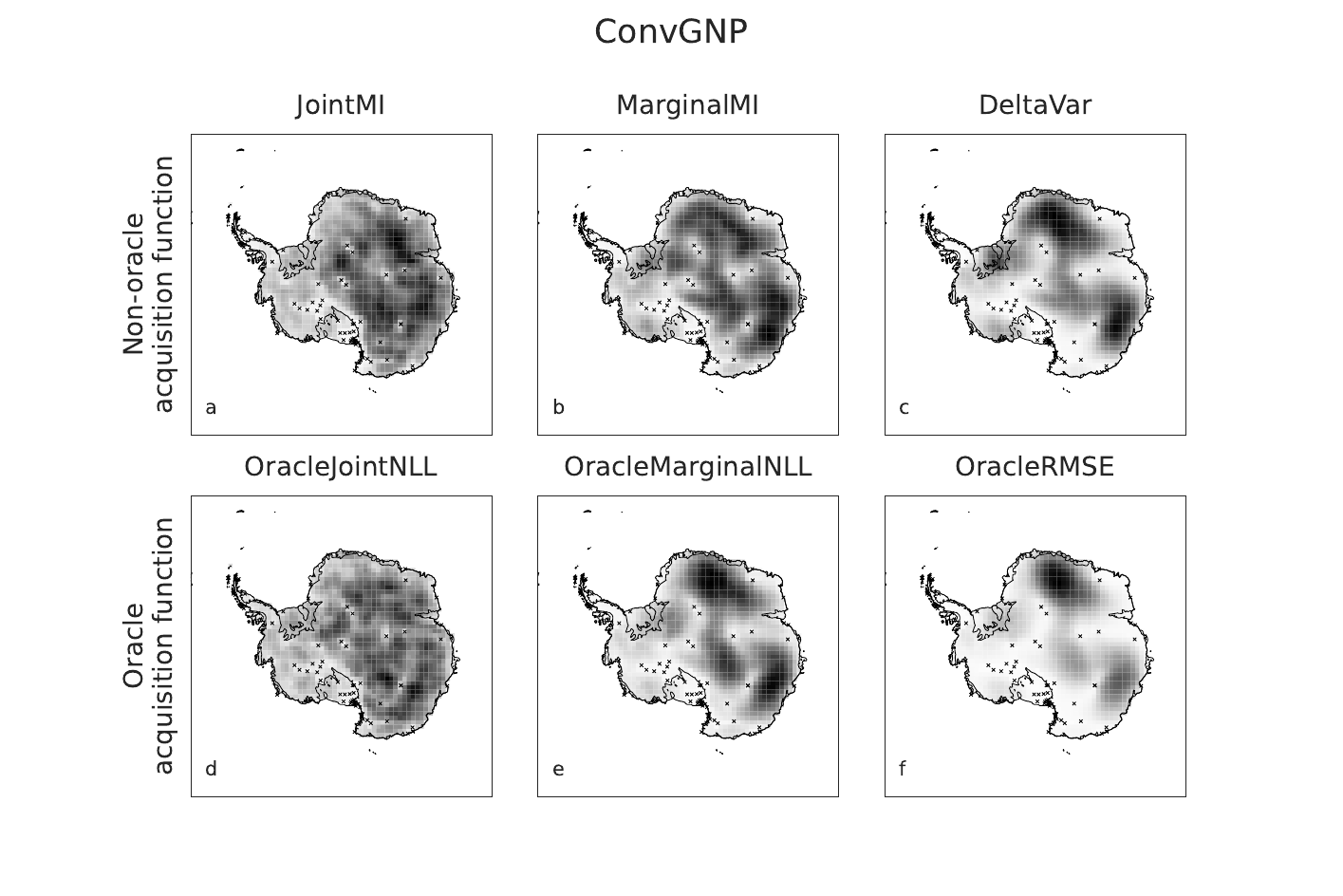}
    \caption{Non-oracle and oracle acquisition functions for the ConvGNP.}
    \label{fig:apx.oracle_convgnp}
\end{figure}

\begin{figure}[t]
    \centering
    \includegraphics[trim=1.2cm 1cm 1.2cm 0cm,clip,scale=1.1]{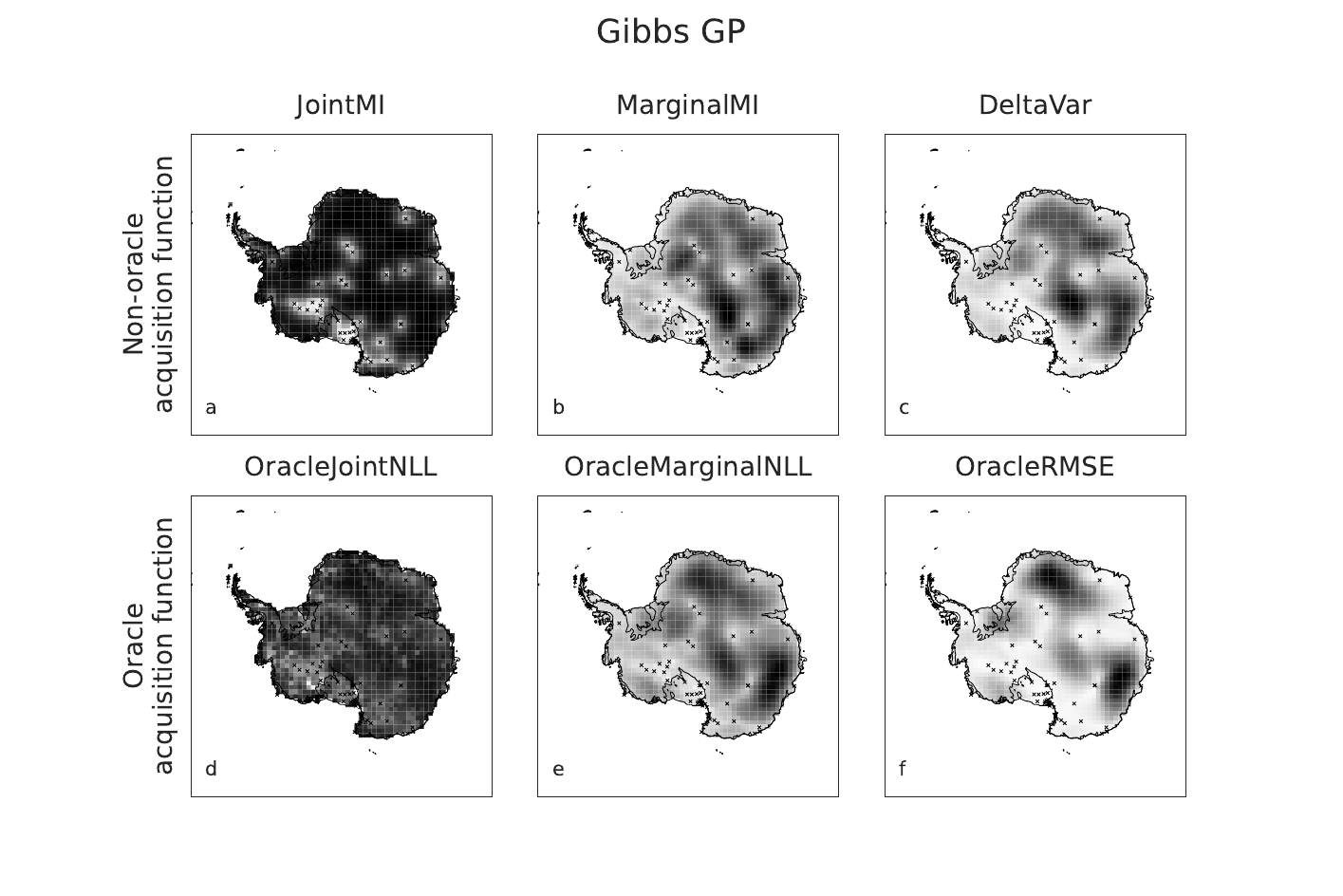}
    \caption{Non-oracle and oracle acquisition functions for the Gibbs GP.}
    \label{fig:apx.oracle_gibbsp}
\end{figure}

\begin{figure}[t]
    \centering
    \includegraphics[trim=1.2cm 1cm 1.2cm 0cm,clip,scale=1.1]{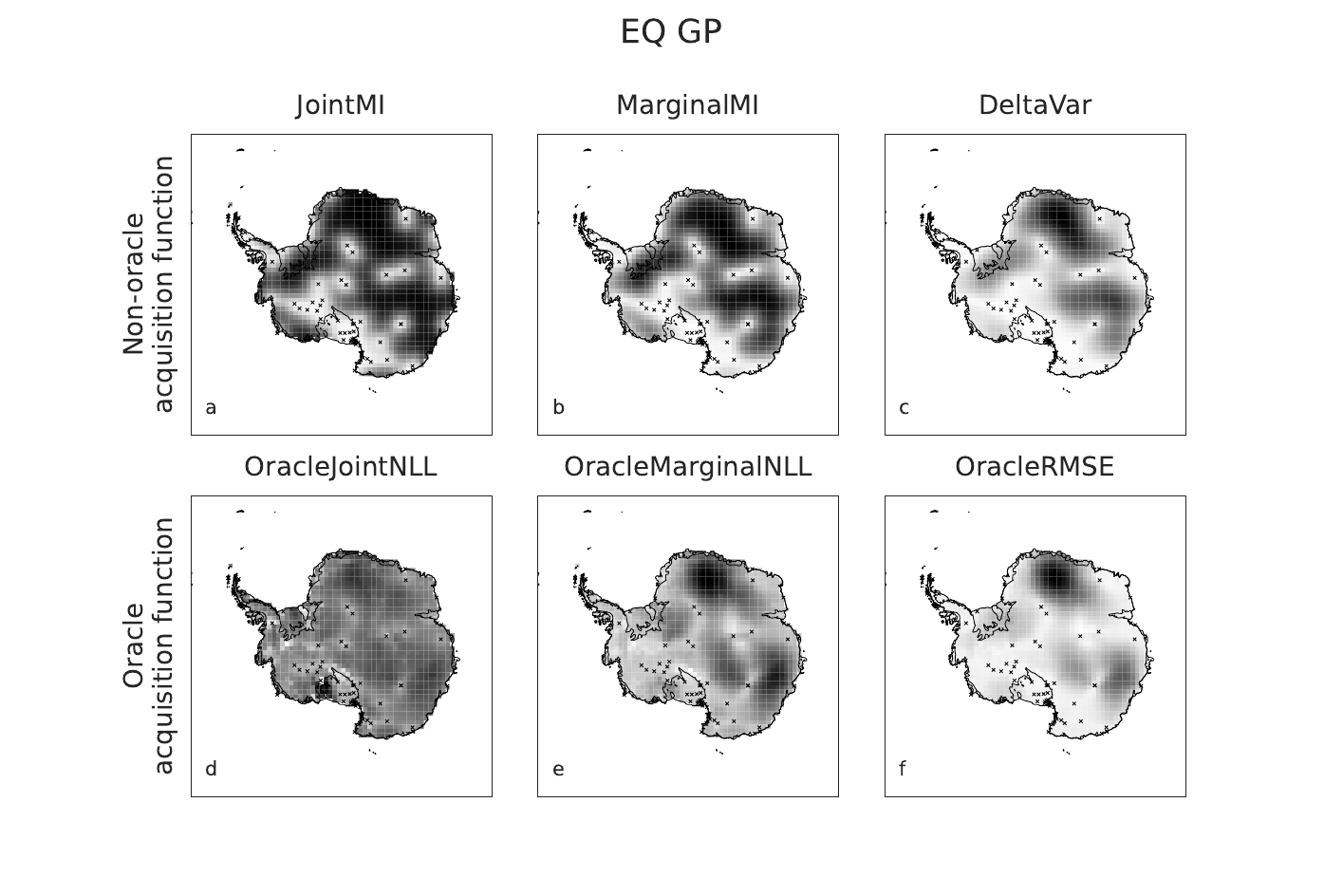}
    \caption{Non-oracle and oracle acquisition functions for the EQ GP.}
    \label{fig:apx.oracle_eq}
\end{figure}

\begin{figure}[t]
    \centering
    \includegraphics[scale=1]{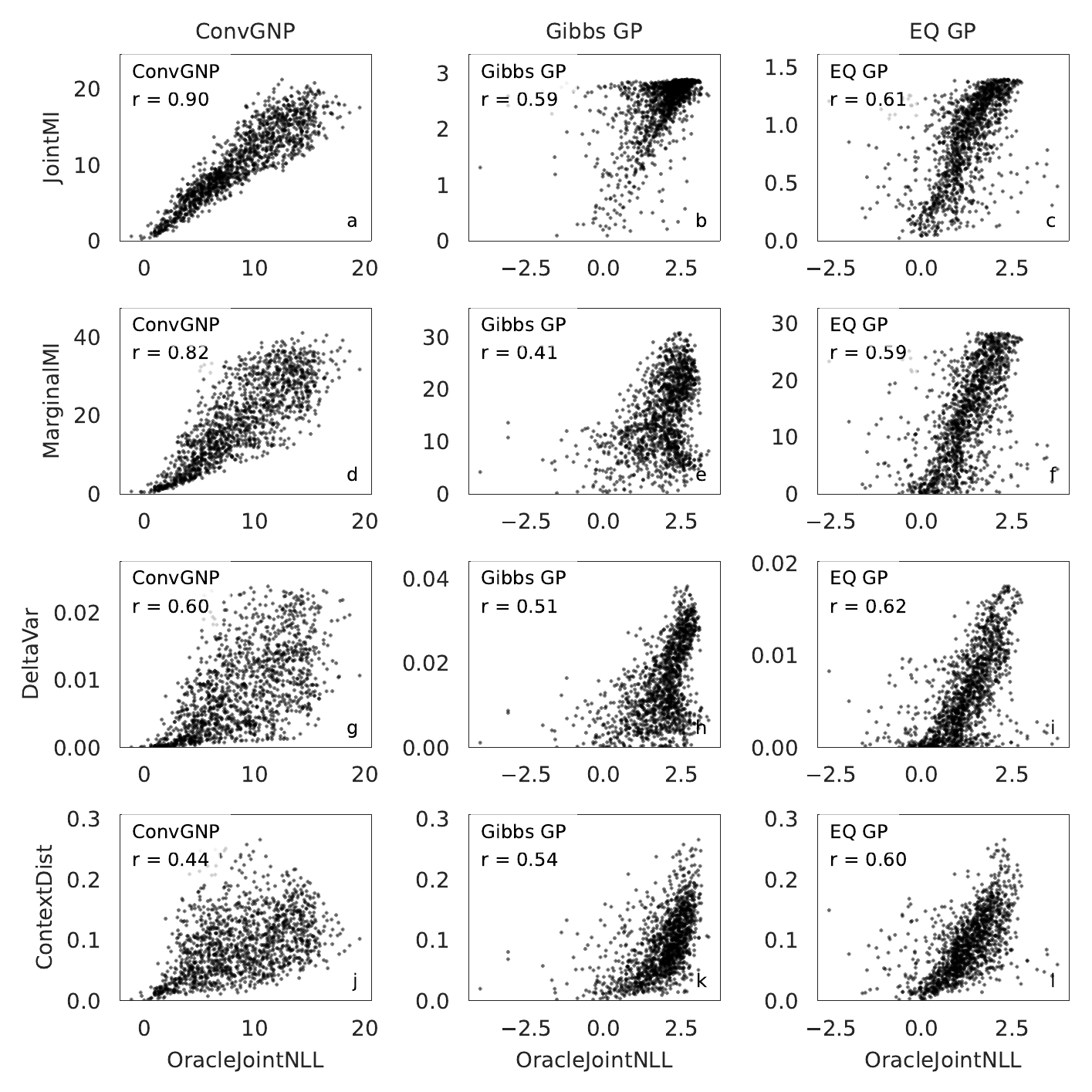}
    \caption{Scatter plots and correlations between non-oracle acquisition functions and \texttt{OracleJointNLL}.}
    \label{fig:apx.oracle_corr_nll}
\end{figure}

\begin{figure}[t]
    \centering
    \includegraphics[scale=1]{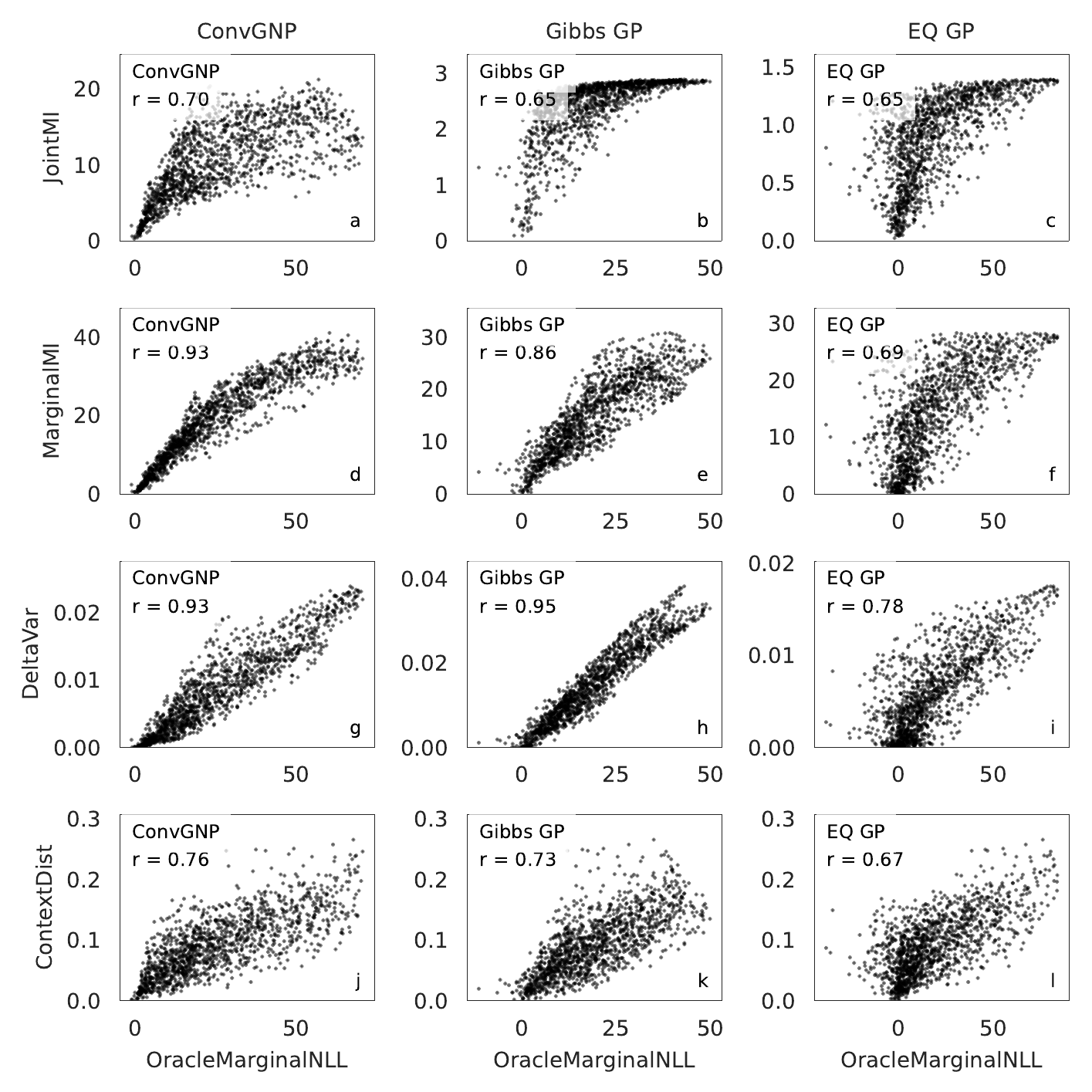}
    \caption{Scatter plots and correlations between non-oracle acquisition functions and \texttt{OracleMarginalNLL}.}
    \label{fig:apx.oracle_corr_marginal_nll}
\end{figure}

\begin{figure}[t]
    \centering
    \includegraphics[scale=1]{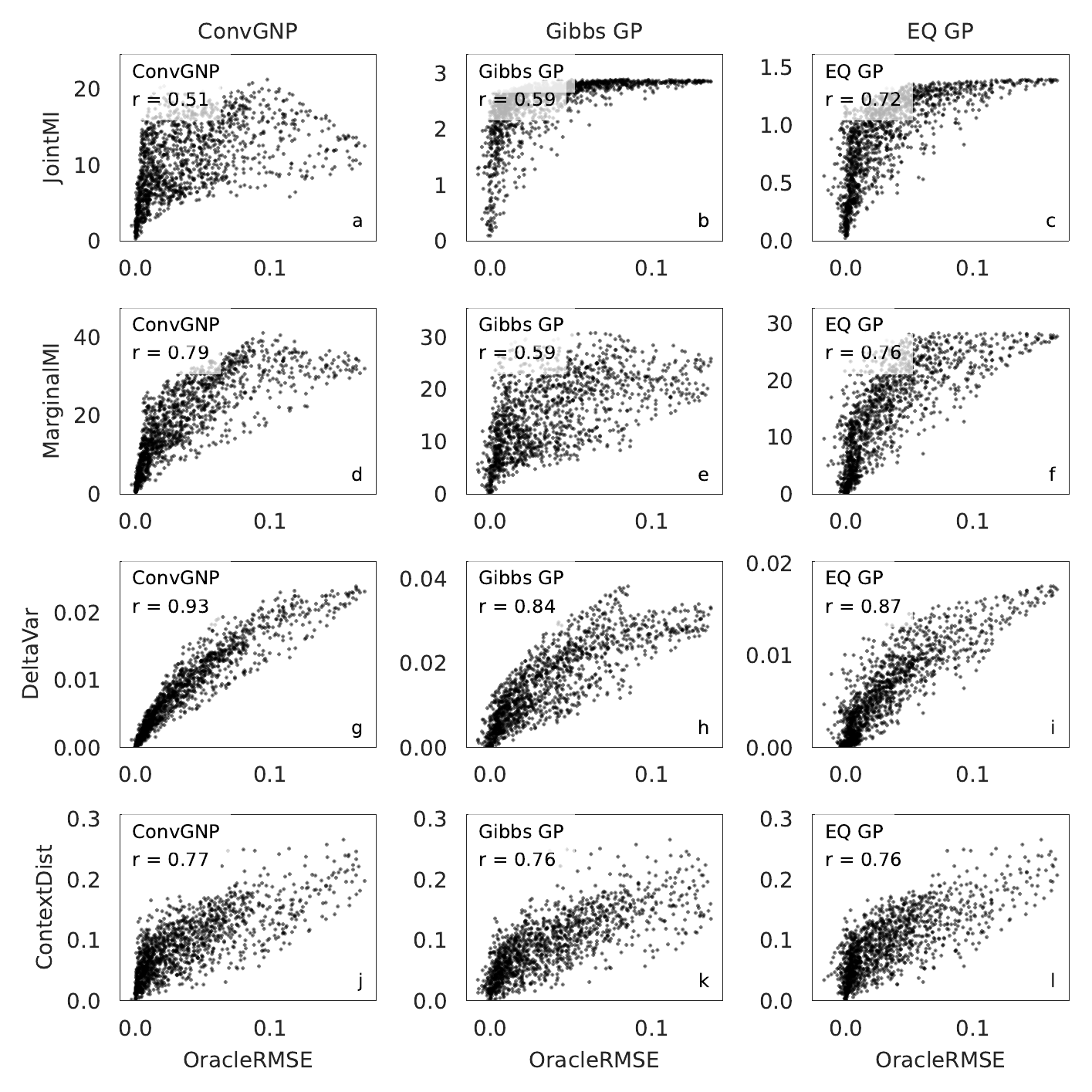}
    \caption{Scatter plots and correlations between non-oracle acquisition functions and \texttt{OracleRMSE}.}
    \label{fig:apx.oracle_corr_rmse}
\end{figure}

Here we provide more detailed plots from the oracle acquisition function experiment described in \Cref{section.oracle_acquisition_functions}.
Heatmaps of the temporally-averaged $\alpha(\bm{x}\us{(s)}_i)$ acquisition functions for the non-oracle and oracle acquisition functions for the ConvGNP, Gibbs GP, and EQ GP are shown in \Cref{fig:apx.oracle_convgnp}, \Cref{fig:apx.oracle_gibbsp}, and \Cref{fig:apx.oracle_eq}, respectively.
\Cref{fig:apx.oracle_corr_nll}, \Cref{fig:apx.oracle_corr_marginal_nll}, and \Cref{fig:apx.oracle_corr_rmse} show scatter plots of non-oracle acquisition functions against \texttt{OracleJointNLL}, \texttt{OracleMarginalNLL}, and \texttt{OracleRMSE}, respectively.

We repeat the Pearson correlation analysis of \Cref{fig:oracle_correlation_results} but using a correlation coefficient specifically suited to rankings, the Kendall rank correlation coefficient:
\begin{equation}\label{eq:apx.kendall}
    \kappa = \frac{1}{N_\text{pairs}}\sum_{i<j}\text{sgn}(\alpha_i - \alpha_j)\text{sgn}(\alpha_{\text{oracle},i} - \alpha_{\text{oracle},j}),
\end{equation}
\noindent where $N_\text{pairs}=S(S-1)/2$ is the total number of pairs, $\alpha_i = \alpha(x^{(s)}_i)$, and $\text{sgn}$ is the sign function which is $+1$ if the argument is positive and $-1$ if the argument is negative..
This loops over all pairs of $(\alpha_i, \alpha_{\text{oracle}, i})$ and $(\alpha_j, \alpha_{\text{oracle}, j})$, checking whether the $\alpha$ values are ranked in the same order as the $\alpha_\text{oracle}$ values.
If so, the pair is `concordant' and contributes a $+1$ to the sum in \Cref{eq:apx.kendall}.
Otherwise, it is `discordant' and contributes a $-1$.
Defining the total number of concordant pairs as $N_\text{con}$, Equation~\ref{eq:apx.kendall} can be rewritten as:
\begin{align}
    \kappa &= \frac{1}{N_\text{pairs}}\big(N_\text{con} - (N_\text{pairs} - N_\text{con})\big), \\ 
     &= 2 \times \frac{N_\text{con}}{N_\text{pairs}} - 1, \label{eq:apx.kendall2}
\end{align}
\noindent which we see as the fraction of pairs that are concordant, $N_\text{con}/N_\text{pairs}$, normalised to lie in $(-1, 1)$.

\begin{figure}[h]
    \centering
    \includegraphics[scale=0.85]{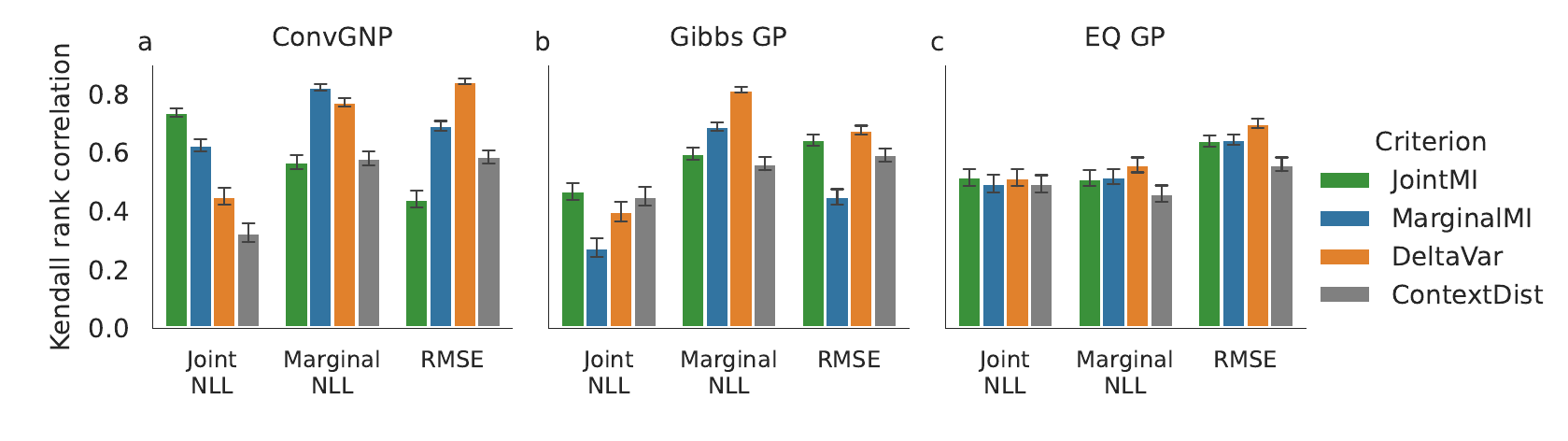}
    \caption{\textbf{The ConvGNP can reliably rank the value of new observations}. Kendall rank correlation coefficient $\kappa$ between model-based and oracle acquisition functions. Error bars indicate the 95\% percentile interval over 5000 bootstrapped correlation values by resampling the 1365 pairs of points with replacement, measuring how spatially consistent $\kappa$ is across space.} 
    \label{fig:apx.oracle_kendalltau_results}
\end{figure}

Identical rankings yield $\kappa=1$, exactly opposite rankings yield $\kappa=-1$, and if the two rankings are independent the expected value of $\kappa$ is zero.
We computed $\kappa$ in Python using the \texttt{scipy.stats.kendalltau} function.
The results are shown in \Cref{fig:apx.oracle_kendalltau_results}.


\FloatBarrier
\section{Sensor placement toy experiment details}\label{apx.section.placement}

Here we provide more details on the sensor placement toy experiment.

\subsection{Experiment design choices}\label{apx.placement.details}

To emulate a non-uniform, real-world sensor network to be optimised, we initialise $\bm{X}_\tau\us{(c)}$ at the locations of real Antarctic temperature station observations at $\tau=2009/02/15$, and interpolate the gridded ERA5 temperature anomaly at $\bm{X}_\tau\us{(c)}$ to compute $\bm{y}_\tau\us{(c)}$.

The \texttt{JointMI}, \texttt{MarginalMI}, and \texttt{DeltaVar} criteria for the GP models and the \texttt{ContextDist} criterion only depend on the context set locations, $\bm{X}_\tau\us{(c)}$, not the observed values $\bm{y}_\tau\us{(t)}$ (\Cref{apx.section.ydependence}).
Since we have a non-varying $\bm{X}_\tau\us{(c)}$ in this toy experiment, the model-based acquisition functions do not depend on time $\tau$ for the GPs, and so the sensor placements for these criteria can be run using a single task.
In contrast, each oracle acquisition function and the ConvGNP's model-based criteria \emph{do} depend on the observed values $\bm{y}_\tau\us{(t)}$.
For these acquisition functions we compute the average $\alpha(\bm{x}\us{(s)}_i)$ values in Equation~\ref{eq:alpha_avg} using dates in 2014--2017, subsampled by a factor of 14, to yield $J=105$ sensor placement search tasks.
A regular spatial grid is used for the search space $\bm{X}\us{(s)}$, with one query location every \SI{100}{km}.
The $\bm{x}\us{(s)}_i$ were masked out over the ocean to focus on land stations.
The target locations $\bm{X}_\tau\us{(t)}$ were defined on the same grid with points over ocean masked out to focus on predicting land surface temperature.
These choices result in a search size and target set size of $S=N_t=1,365$.
Note, limiting the target set size to $N_t=1,365$ was due to the cubic computational cost of the GP baselines--the ConvGNP could use a much denser target grid due to its linear scaling with number of target points.
For the ConvGNP, sequentially computing one of the acquisition functions over these $J=105$ dates and $S=1,365$ search points (totalling $143,325$ forward passes) took roughly 3 hours on a 32 GB NVIDIA V100 GPU using TensorFlow's eager mode.

The proposed placements $\bm{X}^*$ were assessed by analysing model performance over 243 uniformly spaced dates in 2018--2019 (sampling every 3rd day).
The sensor placement search period aligns with the model validation period, while the sensor placement analysis period aligns with the model test period.




\subsection{Full sensor placement results}\label{apx.section.full_placement_results}

\begin{figure}[t]
    \centering
    \includegraphics[scale=0.95]{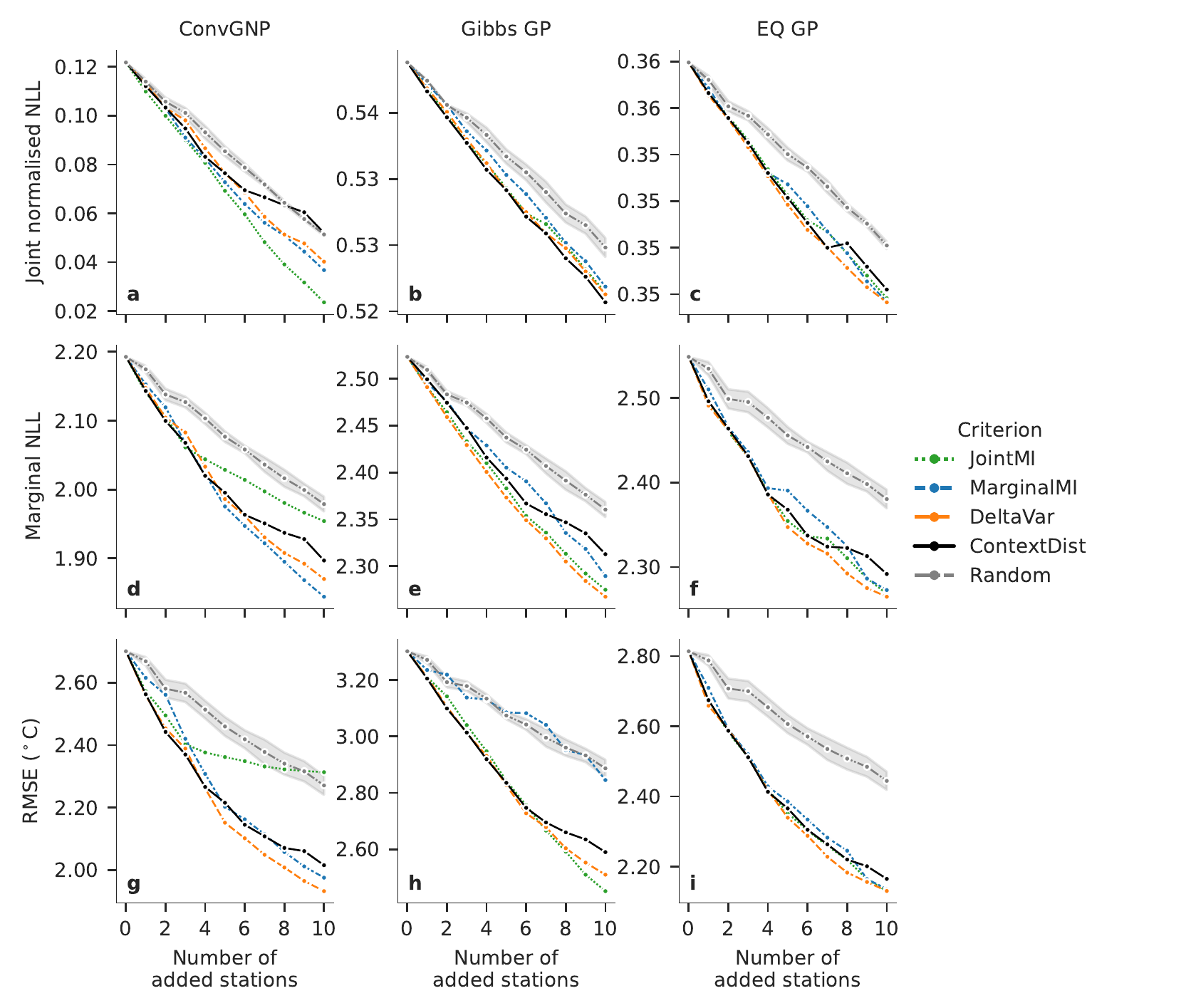}
    \caption{\textbf{Sensor placement results.} Performance metrics on the 2018-2019 sensor placement ERA5 test data versus the number of stations revealed to the models. Results are averaged over tasks with targets defined on a regular grid over Antarctica. For the \texttt{Random} placement criterion, the confidence interval shows the standard error based on 5 random placements. \textbf{a-c}, joint normalised negative log-likelihood (NLL). \textbf{d-f}, mean marginal NLL. \textbf{g-i}, root mean squared error (RMSE).}
    \label{fig:apx.sp_results}
\end{figure}

\begin{figure}[t]
    \centering
    \includegraphics[scale=0.95]{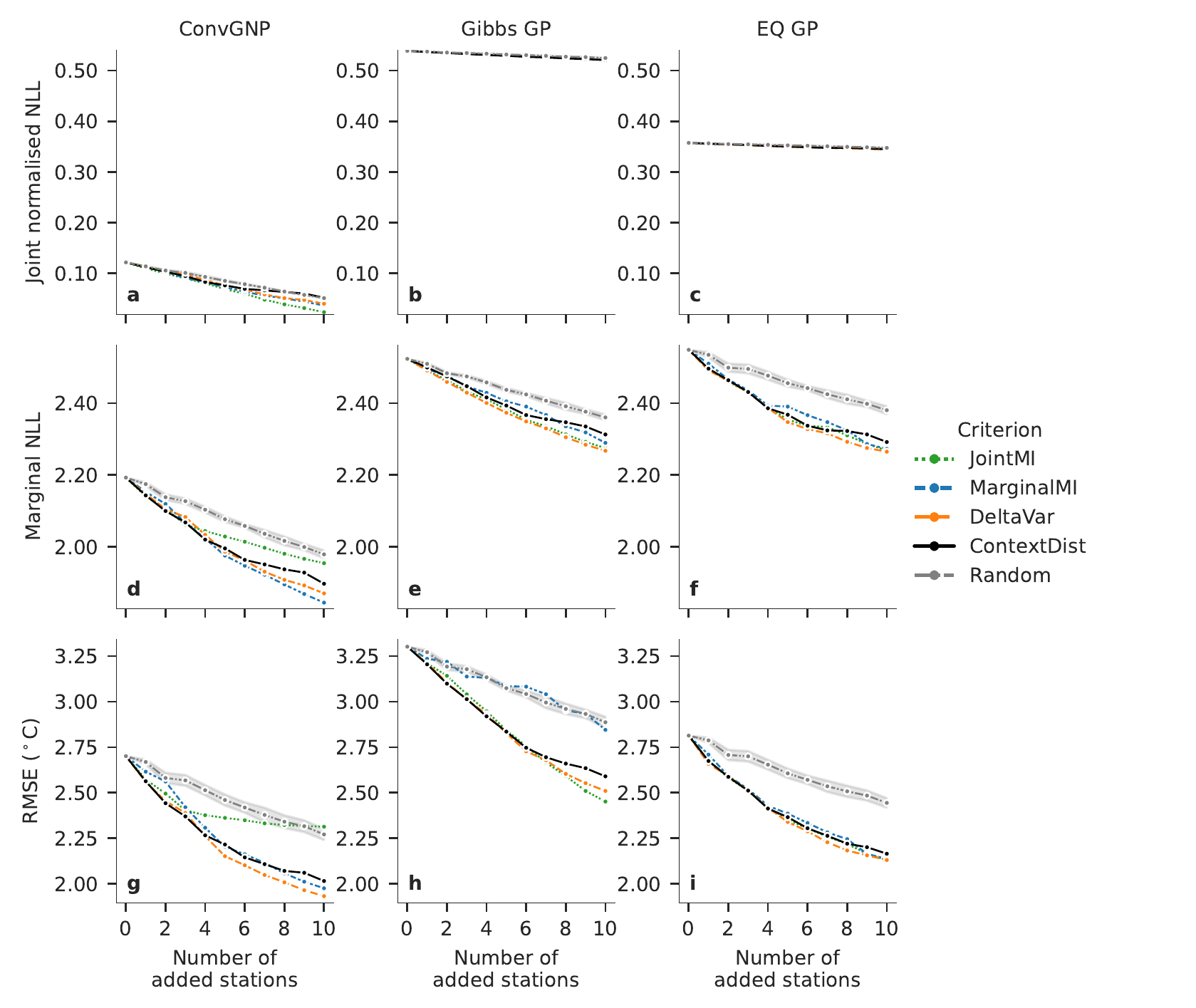}
    \hspace{-1.5cm}
    \caption{\textbf{Sensor placement results with shared y-axes.} Performance metrics on the 2018-2019 sensor placement ERA5 test data versus the number of stations revealed to the models. Placements are revealed in the order of placement for each criterion. Results are averaged over tasks. For the \texttt{Random} placement criterion, the confidence interval shows the standard error based on 5 random placements. \textbf{a-c}, joint normalised negative log-likelihood (NLL). \textbf{d-f}, mean marginal NLL. \textbf{g-i}, root mean squared error (RMSE).}
    \label{fig:apx.sp_results_sharey}
\end{figure}

\Cref{fig:apx.sp_all} plots the $K=10$ proposed sensor placements for each model and placement criterion.
The full breakdown of the sensor placement results for each model, metric, and criterion is shown in \Cref{fig:apx.sp_results}, using independent y-axes to highlight differences in placement criterion performance for a given model.
However, this visually obscures two other differences: initial model performance and the scale of improvement with added stations.
Plotting the results with the y-axes shared across models highlights these differences (\Cref{fig:apx.sp_results_sharey}).

\FloatBarrier

\FloatBarrier
\section{Comparison of the ConvGNP with deep kernel learning}\label{apx.section.dkm}


The ConvGNP is perhaps most comparable to deep kernel learning (DKL; \citealt{wilson2015deep}). 
In DKL, neural networks parameterise a non-stationary covariance function and are trained to optimise a GP prior over the context data.
Then, as with vanilla GPs, standard Bayes' rule conditioning is used with that prior to output posterior GP predictives.
In contrast, the ConvGNP learns to directly output the GP predictive during training.
This allows for outputting GPs that are not in the class of conditioned GP priors, 
which is much more flexible and aids modelling complex environmental data.
However, since robust conditioning is not built in to the ConvGNP, it must learn appropriate conditioning mechanics from the data.
This necessitates a novel training scheme where the model is provided with a range of context scenarios, expanding the training design space and likely making the ConvGNP more data-hungry than DKL.

The ConvGNP scales linearly with the number of context points due to the SetConv encoder and neural network architecture used to output the GP predictive.
Predictions with the ConvGNP's GP predictive are made scalable by directly learning to output a low-rank approximation of the covariance, reducing the computational cost from cubic to linear.
In contrast, DKL methods are by default cubic in the number of context and target points and must use approximate inference on the exact GP to make predictions scalable \citep{wilson2015deep}, resulting in an unknown penalty to prediction quality \citep{wang_exact_2019}.
It is not obvious which is the best approach, although the out-of-the-box nature of the ConvGNP's scalability is convenient from a practitioner's point of view.
Computational cost at inference time is important in the context of environmental applications because observations and target predictions locations may lie on dense grids.

DKL can also be deployed in a meta-learning fashion \citep{patacchiola_bayesian_2020}, which mitigates the risk of overfitting that comes with heavily-parameterised covariance functions \citep{ober_promises_2021}. 
A direct comparison between the meta-learning abilities of the ConvGNP and DKL has not been performed and would be a valuable addition to the literature.

\end{appendix}

\end{document}